\documentclass[remotesensing,article,accept,moreauthors,pdftex,remotesensing]{Definitions/mdpi} 

\usepackage[ruled,linesnumbered]{algorithm2e}
\usepackage{subcaption}
\usepackage{amssymb}
\usepackage{multirow}

\usepackage{upgreek}
\setitemize{parsep=6pt,itemsep=0pt,leftmargin=*,labelsep=5.5mm}
\setenumerate{parsep=6pt,itemsep=0pt,leftmargin=*,labelsep=5.5mm} 
\setlist[description]{itemsep=0mm}

\firstpage{1} 
\pubvolume{12}
\issuenum{14}
\articlenumber{2267}
\pubyear{2020}
\copyrightyear{2020}
\history{Received: 17 June 2020; Accepted: 9 July 2020; Published: 15 July 2020}
\updates{yes} 

\Title{A 
	Soft Computing Approach for Selecting and Combining Spectral Bands}

\Author{Juan F. H. Albarracín $^{1}$\orcidA{}, Rafael S. Oliveira $^{2}$\orcidB{}, Marina Hirota $^{2, 3}$\orcidC{}, Jefersson A. dos Santos $^{4}$\orcidD{} and Ricardo da S. Torres $^{5,}$*\orcidE{}}
\AuthorNames{Juan F. H. Albarracín, Rafael S. Oliveira, Marina Hirota, Jefersson A. dos Santos and Ricardo da S. Torres}

\address{%
$^{1}$ \quad Institute of Computing, University of Campinas, Campinas 13000-000, Brazil; {juan.albarracin@ic.unicamp.br}\\      
$^{2}$ \quad Institute of Biology at University of Campinas, Campinas 13000-000, Brazil; rafaelso@unicamp.br (R.S.O.); marina.hirota@ufsc.br (M.H.)\\    
$^{3}$ \quad Department of Physics, Federal University of Santa Catarina, Florian\'opolis 88040-900, Brazil\\  
$^{4}$ \quad Department of Computer Science at Universidade Federal de Minas Gerais, Belo Horizonte 31270-901, Brazil; {jefersson@dcc.ufmg.br}\\
$^{5}$ \quad Department of ICT and Natural Sciences at Norwegian University of Science and Technology (NTNU), 6009~{\AA}lesund, Norway}

\corres{Correspondence: {ricardo.torres@ntnu.no}; Tel.: {+4792219307}}

\abstract{We introduce a soft computing approach for automatically selecting and combining indices from remote sensing multispectral images that can be used for classification tasks.
The proposed approach is based on a Genetic-Programming (GP) framework, a technique successfully used in a wide variety of optimization problems.
Through GP, it is possible to learn indices that maximize the separability of samples from two different classes. Once the indices specialized for all the pairs of classes are obtained, they are used in pixelwise classification tasks.
We used the GP-based solution to evaluate complex classification problems, such as those that are related to the discrimination of vegetation types within and between tropical biomes. {Using time series defined in terms of the learned spectral indices, we show that the GP framework leads to superior results than other indices that are used to discriminate and classify tropical biomes.}}


\begin{document}



\section{Introduction}
\label{sec:introduction}

Remote sensing is important for monitoring and modeling vegetation dynamics and distribution in large-scale ecological studies. Most of the studies applying these techniques use multispectral indices that are based on a ratio (or some other simple mathematical relation) of the reflectance at two or more wavelengths~\cite{Sims2003:RSE,Xue2017JournalofSensors}.
These indices allow for complex data analyses, such as better detecting and visualizing specific targets, like vegetation vigour~\cite{Ceccato2002:RSE,ZHANG2015157}, water bodies~\cite{Gao1996:RSE}, and~deforestation patterns~\cite{SCHULTZ2016318}, since they make explicit complex interactions between the bands that cannot be evidenced~individually.

The most commonly used indices, such as the Normalized Difference Vegetation Index (NDVI), give a broad and general view of many observable phenomena.
Although these traditional measures are still widely used, they present limitations, among~which we may point out the saturation of the NDVI for closed canopies, preventing the identification of vegetation heterogeneity.
As pointed by Verstraete \& Pinty~\cite{Verstraete1996:TGRS}, spectral indices are very sensitive to the desired information and unresponsive to perturbing factors (such as soil color changes, relief, and~atmospheric effects). 
Thus, they argue that more suitable indices should be designed for specific applications and particular instruments, as~both the desired signal and the perturbing factors vary spectrally, and~the instruments themselves only provide data for certain spectral bands.
In this sense, most of the studies have focused on very specific applications or situations~\cite{885197}, i.e.,~most of them are very sensitive to new data and, thus, are~not easily adaptable to different and possibly new applications or new datasets that are produced from diverse~sensors.

Additionally, if~the specific bands that are used for computing some indices are not available, they~simply cannot be processed (e.g., the~NDVI cannot be calculated from RGB images). \mbox{Finally, most~of} the existing indices ignore available bands in multispectral images. In~particular, land~cover analysis is highly dependent on the sensor used for data acquisition~\cite{doi:10.1080/01431160600746456}. The~ever growing variety of sensors with different technical specifications, such as spatial/spectral resolutions or acquisition protocols, may lead to low performance of traditional spectral indices, as~they may not accurately reflect the same properties related to data obtained from different sources~\cite{Xue2017JournalofSensors}. In~all of those situations, the~use of techniques that can learn existing patterns of interest from data can lead to more effective indices that are tailored to specific classification-dependent~applications.

In this paper, we address those shortcomings by proposing a methodology that employs a Genetic-Programming (GP)-based framework for learning spectral indices that are specialized in discriminating pixels that belong to different classes. GP~\cite{Koza:1992:GPP:138936} is an evolution-based method that models candidate solutions for a problem as individuals of a population, and~makes them evolve through generations. The~objective is to find the best solutions for the target problem as the best individuals of the evolved~population.

GP is widely known because of its capacity of effectively finding complex solutions to a great variety of problems, and~because of its straightforward ability for parallel search~\cite{Koza:1992:GPP:138936}.
Perhaps, the~most important property of GP is the interpretability of the solutions that are learned trough it, since it provides either math formulas or algorithms in a language that is predefined by the user.
This property makes GP more suitable for spectral index learning than other evolutionary algorithms, because~it can be designed to output formulas whose terms are spectral bands, combined through a set of operations that are defined by the user.
The structure of the formula learned with GP can be further analyzed to determine the contribution of the spectral bands, and~then induce high-level knowledge about the objects of study. The~capacity of GP in solving a wide variety of complex problems is evidenced in a large volume of works that has successfully applied GP in uncountable domains~\cite{Taleby2019, Espejo2010, Nguyen2017, Khan2019}.

Band selection and combination methods are important in multispectral scenery classification due to the well-known problem of the curse of dimensionality. {The GP algorithm naturally selects and combines bands in a simultaneous fashion, as~individuals in the population evolve. Hence, our method can be framed as a feature selection and extraction pipeline} (rather than a merely band selection protocol) that projects multispectral images into a one-band feature space, yielded by the learned vegetation index. Even in the multispectral scenario, in~which the few quantity of spectral bands suggests a lower risk of poor performance due to the curse of dimensionality, the~selection of bands prevents classification flaws due to noisy~bands.

We performed experiments to confirm our hypothesis related to the ability of Genetic-Programming-based vegetation indices (GPVI) for representing the canopy structural gradient between regions that can be either forests or savannas biomes, and~the power of such index in extrapolating the classification of forest and savanna areas on a pixel basis for tropical South America, over~time.
{We collected data from four multispectral sensors: Landsat 5 TM, Landsat 7 ETM+, MODIS~MOD09A1.006 (Terra), and~MODIS MYD09A1.006 (Aqua), forming monthly temporal series of the period between 2000 and 2016.}
Biomes are recognized as vegetation units of a particular physiognomy, which is the characteristic spatial arrangement and dominance of plant life forms that explain the vegetation general appearance~\cite{woodward2001,woodward2009}. In~summary, we target four research questions:

\begin{enumerate}[label={Q\arabic*.}, ref={Q\arabic*}]
\item\label{rq1} Are GP-based indices more effective than traditional indices in biome type classification tasks?

\item \label{rq2} Would GP-based indices yield effective results when used in more complex classification problems such as those related to the discrimination of different vegetation types within the same biome?

\item\label{rq3} Are GP-based indices suitable for time-series-based classification tasks?

\item\label{rq4} What are the most frequently selected bands by using the GP-based discovery process?

\end{enumerate}

It is important to state that this framework is not proposed to be a new land cover classification scheme. It is intended as a generic framework for guiding the identification of suitable combinations of spectral bands in remote sensing imagery, which may be used later in classification systems. To~answer~\ref{rq1}, we compare classification performance on multispectral pixels projected by GPVI with respect to the ones projected by the widely used Normalized Difference Vegetation Index (NDVI) and Enhanced Vegetation Index (EVI). Furthermore, we handled two finer-scale classification problems, based on the discrimination of subcategories of savannas and forests with respect to its physiognomical structure and, thus providing an answer to~\ref{rq2}. Biomes are often subdivided into structural, compositional and/or functional subcategories (hereafter called ``vegetation types''), according to plant  phenological patterns (canopy leaf production and loss) or tree cover variation (canopy height and closure), among~others. Analysis by vegetation type yields two sub-classes for forest (evergreen forests vs. semi-deciduous forests) and savannas (typical savannas vs. forested savannas). ({{In the context of this paper}, we use the term forested savanna to refer to a tropical savanna biome which structurally resembles forests~\cite{Ratnan2011GEB}.}) To address~\ref{rq3}, taking advantage of the available temporal data, we perform experiments for whole-time-series classification and analysis, yielded by GPVI and the traditional indices. To~address~\ref{rq4}, we analyze the structure of the learned spectral indices to determine the contribution of each band for~classification.  

\section{Related~Work}
\label{sec:related_work}

GP and other bio-inspired strategies have been used to develop spectral indices in a wide range of applications~\cite{Su2016:JSTARS,Papa2016:JSTARS, Fonlupt2000}.
Chion~et~al.~\cite{Chion2008} proposed the genetic programming-spectral vegetation index (GP-SVI), a~method that evolves a regression model to describe the nitrogen level in vegetation.
A~very similar work by Puente~et~al.~\cite{Puente2011} introduced the Genetic Programming vegetation index (GPSVI) to estimate the vegetation cover factor in soils to assess erosion.
Ceccato~et~al. proposed a new methodology to create a spectral index to retrieve vegetation water content from visible data~\cite{Ceccato2002:RSE}. Gerstmann~et~al.~\cite{Gerstmann2016115}, in~turn, introduced a method to find a general-form normalized difference index to separate cereal crops, through an exhaustive search in a set of possible permutations of bands and constants used in a general formula and measures effect size to determine class~separability.

Spectral indices have demonstrated an ability to separate different classes of woody vegetation~\cite{rs9010005}. However, efforts that apply spectral indices for classification purposes normally rely on time series~\cite{5509070, Low201591,Prishchepov2012195}. A~single value in time of the indices about an object of interest is not enough to discriminate it. This is particularly true in vegetation analysis since an object of interest can change its configuration seasonally.
The possibility of novel indices that encode discriminative information invariant through time is open and could be achieved with our proposed technique.
Balzarolo~et~al.~\cite{Balzarolo2016:RSE} showed that, while using the NDVI index, there is a high correlation between the starting day of the growing season observed with MODIS data, and~in-situ observations.
Wang~et~al.~\cite{Wang2017:RSE} developed a three-band vegetation index to minimize the snow-melt effect in monitoring spring~phenology.

GP has also been used for classification. Ross~et~al. presented a method for mineral classification using three classes~\cite{Ross2005}.
Rauss~et~al.~\cite{rauss2000classification} proposed evolving an index that returns values greater than $0$ when there is grass in the image, and~values smaller than $0$, otherwise. 
In general, approaches for classification are applied in very close configurations like few classes or isolated binary cases.
Regarding function learning for general purposes, GP has also been used to learn formulas to combine time series similarity functions~\cite{Almeida2017, Menini2019}, or~formulas that extract relevant information from similarity measures obtained throughout the user relevance feedback iterations~\cite{Calumby2014}, or~textual sources~\cite{Saraiva2013}.

Different from the previous attempts, we propose a method that relies on how data are distributed in the space yielded by candidate indices, instead of driving the learning process exclusively to find indices that satisfy specific purposes (e.g., regression, classification).
Albarracín~et~al.~\cite{hernandez2016} explored the idea of learning spectral indices for classification tasks. Different from that work, here we conduct a comprehensive formal description of the GP-based index discovery framework.
The objective is to guide researchers and developers in the creation of novel realizations and extensions of the proposed approach for managing multispectral data for different applications.
To the best of our knowledge, this is one of the first works to describe experiments while using GP-based indices in such problems. We also add further analysis about the relevance of bands found in the learned vegetation indices. This may provide insights about the most important environmental processes that characterize each biome evaluated here. Finally, we introduce an optimized fitness function that is designed for a binary classification~problem.

\section{Learning Indices Based on Genetic Programming for Classification~Tasks}

\begin{figure}[H]
\centering
\includegraphics[width=1\textwidth]{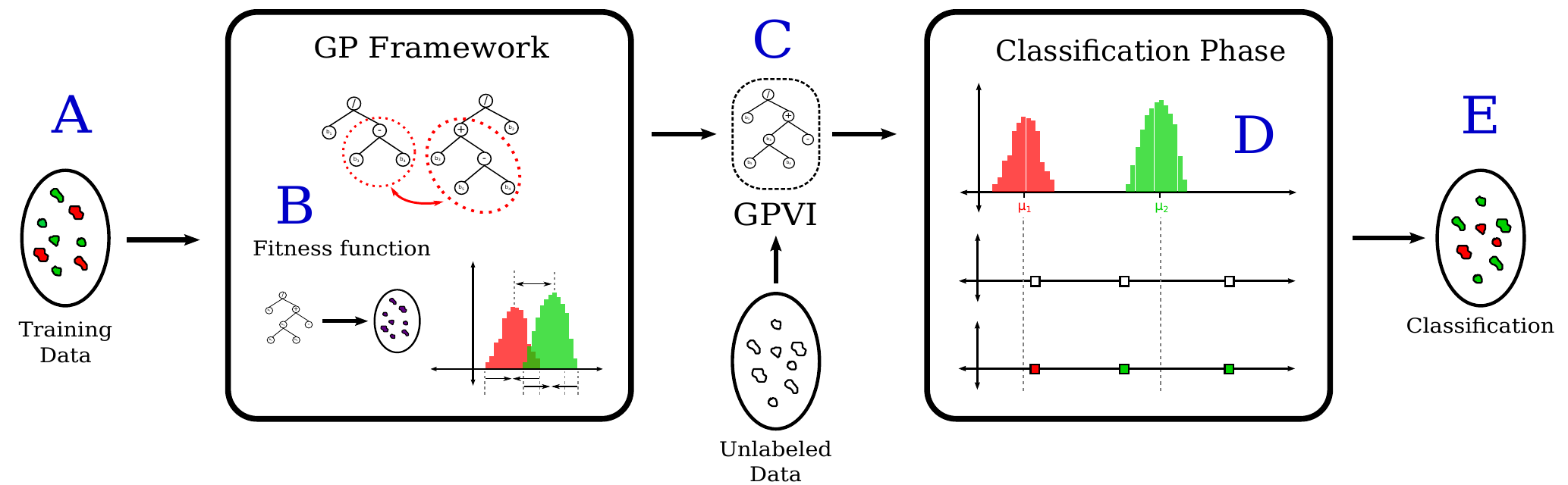}
\caption{Genetic-Programming-based vegetation indices (GPVI) learning process and posterior classification framework for pixelwise discrimination. The~labeled training set is represented by the set defined by part ``A''. This ground truth is then used in the GP-based index discovery process in part ``B''---see Sections~\ref{sec:GP} and~\ref{sec:GP-index}. Classification is performed based on the best GP individual (GPVI) found---part ``C''. This individual is used to define a score for unlabeled samples and to define the index distributions per class. Finally, unlabeled samples are assigned to the class whose distribution is the closest---part~``E''.}
\label{fig:pipeline}
\end{figure}

In this section, we introduce the proposed GP-based index discovery framework and discuss its use in classification problems. First, we provide background on GP concepts (Section~\ref{sec:GP}). Subsequently, we explain how GP is used to find suitable indices (Section~\ref{sec:GP-index}). Finally, we discuss the use of GP-based indices in classification settings (Section~\ref{sec:GP-index-classification}). The~whole pipeline is depicted in Figure~\ref{fig:pipeline}.

\unskip

\subsection{Background on Genetic~Programming}
\label{sec:GP}

Genetic Programming (GP) is an artificial intelligence framework that is used for addressing optimization problems, based on the theory of evolution~\cite{Koza:1992:GPP:138936}.
GP represents candidate solutions to a problem as individuals within a population, which evolves over generations. The~objective is to search for the best solutions (i.e., best individuals of the population) for the target problem, according to the Darwinian principle of survival of the fittest. In~order to determine how fit an individual is to be selected, i.e.,~how good it solves the problem, a~criterion must be defined in terms of a fitness function. This function has as input an individual, and~outputs a score that allows for comparison of all the individuals in the population. This score determines the chance of individuals to be~selected.

\begin{figure}[H]
\centering
\includegraphics[width=.6\columnwidth]{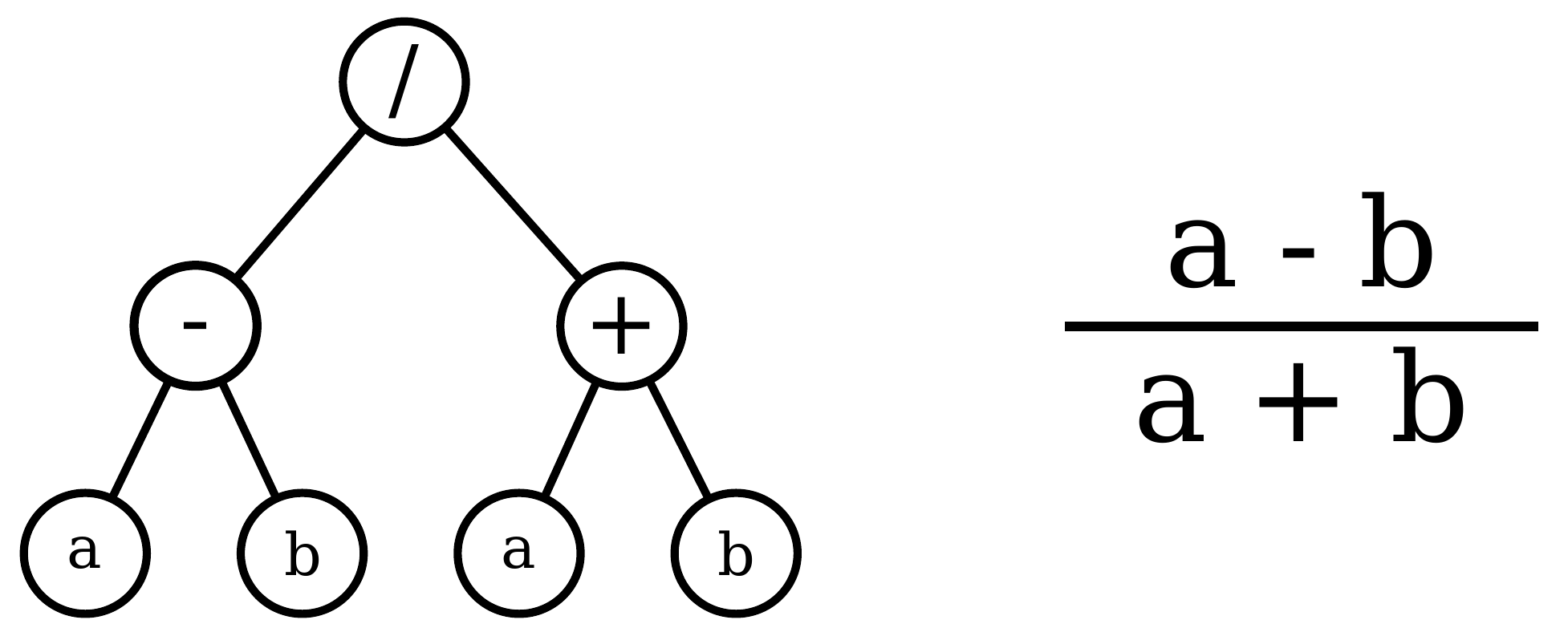}
\caption{Genotype (\textbf{left}) and phenotype (\textbf{right}) of a mathematical~formula.}
\label{fig:gp_individual}
\end{figure}

A typical representation of a GP individual relies on the use of a tree data structure, where~leaf nodes (also known as terminals) represent operands and internal nodes encode functions, usually~mathematical operators used to combine terminals. Figure~\ref{fig:gp_individual} illustrates a typical GP tree representation, which encodes the function $f(a,b)= (a - b) \div (a + b)$, where $a$ and $b$ represent two terminals (variables or constants) and $+$, $-$, and~$\div$ are the internal~nodes.

\begin{figure}[H]
\centering
\includegraphics[width=.757\columnwidth]{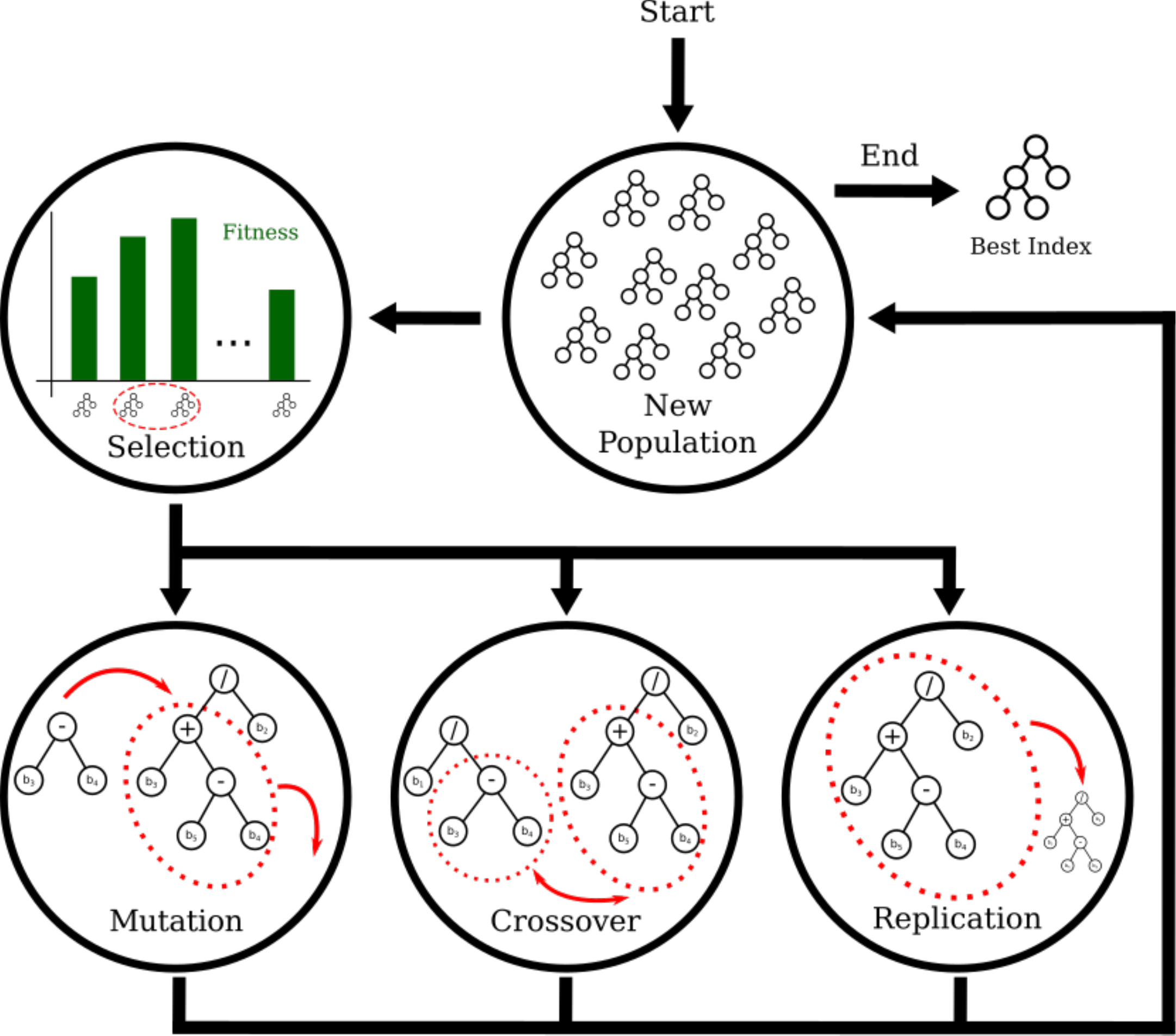}
\caption{The general genetic programming~process.}
\label{fig:gp}
\end{figure}

Figure~\ref{fig:gp} illustrates the typical evolutionary steps considered in the GP-based discovery process, which are outlined in Algorithm~\ref{alg:gp}.
GP starts with a randomly generated population of individuals (Line~\ref{algline:init}). The~fitness of each individual is computed (Line~\ref{algline:init_eval}) and a selection strategy is employed (Line~\ref{algline:init_select}) to determine which ones will be subjected to genetic operations. Typical operations include replication, mutation, and~crossover. Each genetic operation has an associated occurrence rate, that~determines how likely it is to be applied (Line~\ref{algline:select_op}).
In replication (Line~\ref{linalg:replication}), individuals are selected to be placed in the population of the next generation, based on their scores. The~diversity of candidate solutions in the population is maintained by mutation (Line~\ref{linalg:mutation}), which is usually implemented as the substitution of a randomly selected node (operators or even a complete subtree can be selected to be changed) by a randomly generated new subtree. Finally, crossover (Line~\ref{linalg:crossover}) is usually implemented as the exchange of subtrees from two other individuals (parents), which are selected based on their scores. The~process of selecting individuals and applying genetic operators to form a new population of individuals is known as a generation (Line~\ref{linalg:generation}). The~algorithm keeps iterating through generations until a stop condition is reached (Line~\ref{linalg:stop}), e.g.,~a pre-defined maximum number of generations. The~best individual found so far is always preserved (Line~\ref{linalg:best}), to~be returned by the algorithm at the~end.

\begin{algorithm}
\small
\SetKwInOut{Input}{input}
\SetKwInOut{Output}{output}
\SetAlgoLined
\Input{$pop\_size$, $nodes$, $P_{crossover}$, $P_{mutation}$, $P_{replication}$}
\Output{$S_{best}$}

$Population \longleftarrow InitializePopulation(pop\_size, nodes)$\\
\label{algline:init}
$EvaluatePopulation(Population)$\\
\label{algline:init_eval}
$S_{best} \longleftarrow GetBestSolution(Population)$
\label{algline:init_select}

 \While{ $\neg$StopCondition()}{
 \label{linalg:stop}
  $Children \longleftarrow \emptyset$\\
  \While{ $Size(Children) < population\_size$ }{
  $Operator \longleftarrow SelectOperator(P_{crossover}, P_{mutation}, P_{replication})$\\
  \label{algline:select_op}
  \uIf{$Operator \equiv CrossoverOperator$}{
  \label{linalg:crossover}
   $Parent_1, Parent_2 \longleftarrow SelectParents(Population, pop\_size)$\\
   \label{linalg:sel1}
   $Child_1, Child_2 \longleftarrow Crossover(Parent_1, Parent_2)$\\
   $Children \longleftarrow Children \cup \{Child_1, Child_2\}$
   }
   \uElseIf{$Operator \equiv MutationOperator$}{
   \label{linalg:mutation}
   $Parent_1 \longleftarrow SelectParents(Population, pop\_size)$\\
   \label{linalg:sel2}
   $Child_1 \longleftarrow Mutate(Parent_1)$\\
   $Children \longleftarrow Children \cup \{Child_1\}$
  }
  \uElseIf{$Operator \equiv ReplicationOperator$}{
  \label{linalg:replication}
   $Parent_1 \longleftarrow SelectParents(Population, pop\_size)$\\
   \label{linalg:sel3}
   $Child_1 \longleftarrow Replicate(Parent_1)$\\
   $Children \longleftarrow Children \cup \{Child_1\}$
  }
  }
  $EvaluatePopulation(Children)$\\
  $S_{best} \longleftarrow GetBestSolution(Children, S_{best})$\\
  \label{linalg:best}
  $Population \longleftarrow Children$
  \label{linalg:generation}
 }
 \Return $S_{best}$
 
 \caption{Typical Genetic Programming~algorithm.}
\label{alg:gp}
\end{algorithm}
\vspace{-6pt}
\subsection{Gp-Based Index Discovery~Process}
\label{sec:GP-index}
\vspace{-6pt}
\subsubsection{Problem Definition} We apply the GP framework that is described in Section~\ref{sec:GP} to automatically learn spectral indices from remote sensing image (RSI) data. The~GP-index discovery process relies on the definition of a training set composed of multispectral image pixels, which belong to distinct classes and their corresponding labels (e.g., savannas or forests). The~framework is then used to evolve formulas, which~represent mathematical equations combining the bands to find an index that successfully discriminates pixels belonging to the two~classes.

\subsubsection{Individual Representation} 
\begin{figure}[H]
\centering
\includegraphics[width=0.4\columnwidth]{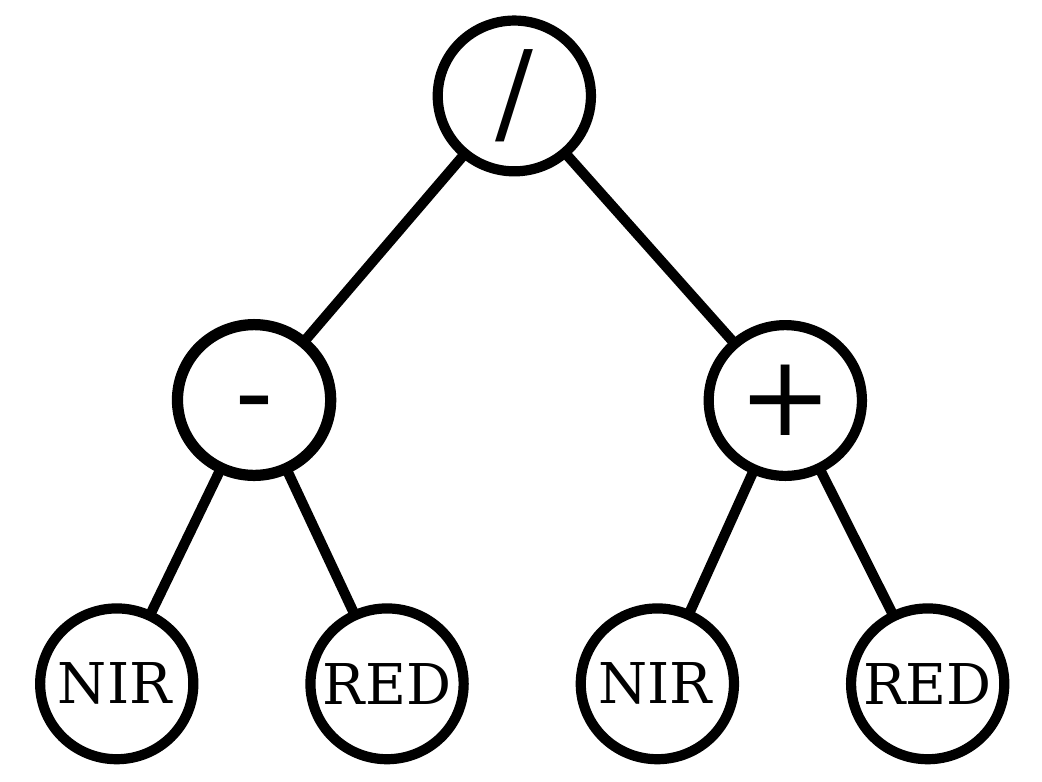}
\caption{Syntax tree representing the~NDVI.}
\label{fig:gp_ndvi}
\end{figure}

In our formulation, the~spectral indices are computationally represented as a GP tree whose leaves are associated with RSI bands and the internal nodes, arithmetic operations. Figure~\ref{fig:gp_ndvi} illustrates a GP individual, encoding the equation used to compute the Normalized Difference Vegetation Index (NDVI), defined in Equation~(\ref{eq:ndvi}).

\subsubsection{Discovery Process} The GP framework starts with a randomly generated population of indices (e.g., $b_1 + b_7 \times b_2$) that evolve as described~above.

\subsubsection{Fitness Computation} We devised a fitness function, which measures how well each GP individual (index) discriminates pixels from different classes. Each candidate index is used to combine bands of the pixels in the training set, associating one scalar with each one of them. With~this information, we obtained index distributions that formed by the training pixels. Because~the multi-band-image pixels used in the training set are labeled (i.e., for~each pixel we know the class it belongs to), it is easy to determine how separated the distributions of indices of different classes are. The~fitness of an individual is then computed based on the inter and intra-class distances in the training set of pixels: those of the same class should have similar index scores and should be far from pixels of the other~class.

Algorithm~\ref{alg:fitness} outlines the fitness computational process. First, we obtain the index distributions for each class (Lines~\ref{linalg:dist_a} and~\ref{linalg:dist_b}), and~we then compute the distance of the means of each distribution, normalized by their standard deviation (Line~\ref{linalg:norm_dist}). Let $\mu_a$ and $\mu_b$ be the mean of all pixel values of, respectively, classes $a$ and $b$, and~let $\sigma_a$ and $\sigma_b$ be their corresponding standard deviations. We defined the score of separability (fitness function) of the two classes as:
\begin{equation}
\label{eq:norm-dist-means}
S = \displaystyle\frac{|\mu_a - \mu_b|}{\max\{\sigma_a,\sigma_b\}}
\end{equation}

The $\max$ function is used instead of the sum or product for the standard deviation values, because~a small standard deviation of one of the classes could compensate the large standard deviation of the other, obtaining high separability values, even if both distributions would happen to overlap. This measure returns real numbers greater or equal than $0$. The~larger the score, the~better the separability. This fitness function is faster than the silhouette score, used in~\cite{hernandez2016}.

\begin{algorithm}
\small
\SetKwInOut{Input}{input}
\SetKw{KwEach}{each}
\SetKw{KwIn}{in}
\SetAlgoLined
\Input{$population, X_{data}, X_{labels}$}

\For{\KwEach $ind$ \KwIn $population$}{
  $X_{ind} \gets ApplyIndex(X_{data}, ind)$ \\
  $X_a \gets SelectDataOfLabel(X_{ind}, a)$ \\
  \label{linalg:dist_a}
  $X_b \gets SelectDataOfLabel(X_{ind}, b)$ \\
  \label{linalg:dist_b}
  $S \gets \displaystyle\frac{|\mu_a - \mu_b|}{\max\{\sigma_a,\sigma_b\}}$ \\
  \label{linalg:norm_dist}
  $AssignFitnessToIndividual(ind, S)$
}
\caption{Fitness~function}
\label{alg:fitness}
\end{algorithm}

\subsubsection{Selection Strategy} The score, associated with each individual by means of the fitness function described above, selects for individuals (see Lines~\ref{linalg:sel1},~\ref{linalg:sel2}, and~\ref{linalg:sel3} of Algorithm~\ref{alg:gp}) and applies the genetic operators on them. The~selection strategy used is called tournament (Algorithm~\ref{alg:tournament}). Given an integer $k$ and the population, we randomly choose $k$ individuals, returning the best of them, according to their~fitness.

\begin{algorithm}
\small
\SetKwInOut{Input}{input}
\SetKwInOut{Output}{output}
\SetKw{KwIn}{in}
\SetAlgoLined
\Input{$population, k$}
\Output{$best$}

$best \gets NIL$

\For{$i = 0$ \KwTo $k$}{
  $i \gets random(1, size(population))$\\
  $ind \gets population\big[i \big]$\\
  \If{$best = NIL \vee getFitness(ind) > getFitness(best)$}{
  $best \gets ind$
  }
}
\Return $best$
\caption{Tournament~selection}
\label{alg:tournament}
\end{algorithm}

\subsection{Using Gp-Based Indices in Classification~Tasks}
\label{sec:GP-index-classification}

After we find a good index, we use it for classification. The~class assignment process is based on the distribution of each class in the index space, through an algorithm known as Nearest Centroid Classifier (NCC)~\cite{introIR} ([Ch.~14]), as~outlined in Algorithm~\ref{alg:ncc}. First, we calculate the index values for all samples (Lines~\ref{linalg:apply_train} and~\ref{linalg:apply_val}). Next, for~the labeled samples, we calculate 
the mean value for each class (Line~\ref{linalg:centroids}) and, for~every validation example, we assign the label of the nearest class centroid (Line~\ref{linalg:find_nc}).

\begin{algorithm}
\small
\SetKwInOut{Input}{input}
\SetKwInOut{Output}{output}
\SetKw{KwEach}{each}
\SetKw{KwIn}{in}
\SetAlgoLined
\Input{$X_{train}, Y_{train}, X_{val}, vi$}
\Output{$Y_{val}$}

$X_{train} \gets ApplyIndex(X_{train}, vi)$ \\
\label{linalg:apply_train}
$X_{val} \gets ApplyIndex(X_{val}, vi)$\\
\label{linalg:apply_val}
$\mu \gets CalculateCentroidsPerClass(X_{train}, Y_{train})$\\
\label{linalg:centroids}
$Y_{val} \gets empty\_list()$

\For{\KwEach $x_i$ \KwIn $X_{val}$}{
  $\mu_i \gets FindNearestCentroidLabel(x_i, \mu)$ \\
  \label{linalg:find_nc}
  $AppendToList(Y_{val}, \mu_i)$
}
\Return $Y_{val}$
\caption{Nearest Centroid~Classifier}
\label{alg:ncc}
\end{algorithm}

\subsection{Computational~Complexity}

The computational complexity of the GP-based index discovery process is $O(G \times N \times E)$, where~$G$ is the number of generations, $N$ is the number of individuals in the population, and~$E$ is the fitness evaluation time. The~calculation time of the fitness function for each individual (see Equation~(\ref{eq:norm-dist-means})) increases linearly with the number of samples used to calculate the distance between the distributions. Thus, $E$ is determined by the complexity of an individual (e.g., its size $S$) and the number of training samples ($n$), which leads to a total time complexity of GP training experiment of $O(G \times N \times S \times n$). Note that this GP learning state is performed once for a given training dataset and an off-line operation. The~application of the learned index on unseen spectral data is performed in constant time, similarly to the traditional vegetation~indices.

\section{Experiments and~Results}
\label{sec:results}

In this section, we separately address the research questions, stated in Section~\ref{sec:introduction}. Therefore, each~question is addressed in a specific subsection, which presents the adopted experimental protocol, used datasets (when necessary), and~achieved~results.

Additionally, we provide a set of experiments that are orthogonal to the ones that involve the data from Tropical South America (\ref{rq1},~\ref{rq2},~\ref{rq3}, and~\ref{rq4}). In~these experiments, we test sensor invariance in GPVI, i.e.,~how robust is our method when tested on different sensors. Therefore, all experiments presented in Sections~\ref{subsec:biome}--\ref{subsec:bands} consider results that are related to the use of two families of sensors: Landsat and~MODIS.

\subsection{Gpvi for Broad-Scale Biome~Classification}
\label{subsec:biome}

In this section, we address~\ref{rq1} by performing experiments for the forest/savanna discrimination problem on raw (non-temporarily ordered) pixels, in~which traditional vegetation indices (e.g., NDVI, EVI, and~EVI2) are widely~used.

\subsubsection{Data Acquisition and~Pre-Processing}
\label{subsubsec:data-tropical}

\begin{figure}[H]
\centering
\includegraphics[width=1\columnwidth]{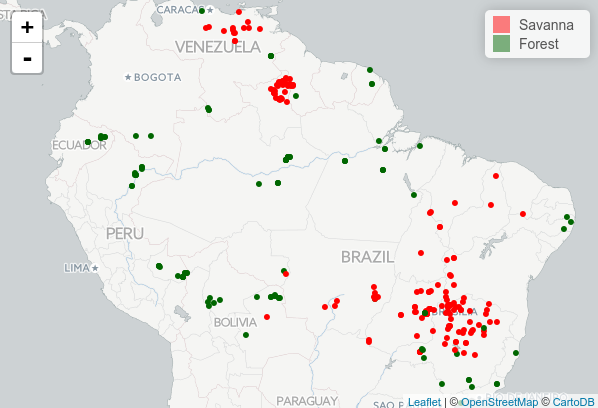}
\caption{Geographical distribution of the areas considered in Tropical South America, corresponding to the data collected for training and validation. Imported form~\cite{Dantas2016}.}
\label{fig:map-f-s}
\end{figure}

The selected study sites are distributed across the region, comprising South American tropical savannas and forests (Figure~\ref{fig:map-f-s}). These locations correspond to 250 m $\times$ 250 m areas with the centroids imported from a subset of the inventory data used by Dantas~et~al.~\cite{Dantas2016}. Each of these points is visually assessed using Google Earth (GE) images from 2016, superimposed with MODIS grid cells of the same size while using Series Views~\cite{Freitas_virtuallaboratory}. The~geographical coordinates used as ground truth are defined in terms of whether the MODIS pixel at which the points fall into are completely filled with the vegetation type consistent with the corresponding labeled biome (i.e., savanna or forest) and vegetation types, as~determined during the field work campaigns. Pixels that are not entirely filled with the corresponding biome type (e.g., due to deforestation and conversion to other land uses) are either replaced by a neighboring pixel fulfilling these criteria or discarded (when a nearby suitable pixel could not be found).

The data were acquired using the Google Earth Engine (GEE) API to download the scenes containing the study areas along the time. Two different data products from different sensors were~downloaded:

{Landsat Surface Reflectance sensors}
	 (Landsat) ({\url{https://www.usgs.gov/products/data-and-tools/gis-data} ({As of  3rd March 2020}))}: we merged data from Landsat 5 Thematic Mapper (TM), and~Landsat 7 Enhanced Thematic Mapper Plus (ETM+), as~suggested by Kovalskyy and Roy~\cite{KOVALSKYY2013280}, in~order to alleviate the lack of data availability due to the Landsat 7 ETM+ scan line failure in~2003. The~GGE API retrieves the data form the United States Geological Survey (USGS), which provides Level-1 terrain-corrected (L1T) products, and~already pre-processes the scenes regarding cloud detection (with the \textit{fmask} algorithm~\cite{ZHU201283}), geo-referencing, and~atmospheric~correction.  

{Moderate Resolution Imaging Spectroradiometer (MODIS)
	:} here, the~data from the satellite products MOD09A1.006 (Terra) ({\url{https://lpdaac.usgs.gov/products/mod09a1v006/} ({As~of  \mbox{3rd March 2020}})}) and MYD09A1.006 (Aqua) ({\url{https://lpdaac.usgs.gov/products/myd09a1v006/} ({As~of  3rd March 2020})}) were also merged. According to the specifications, ({\url{https://nsidc.org/data/modis/terra_aqua_differences} ({As of  24th February 2020}))} the differences between the two satellites do not involve the nature of data collected in the bands considered in this study.
{Data with the blue band greater than 10\% of its maximum value and/or the sensor view zenith angle greater than $32.5^{\circ}$ were excluded, as~discussed by Freitas~et al.~\cite{Freitas_virtuallaboratory}.}

The data are available at 250 m/pixel for {all the bands in the LANDSAT sensors, and~only in bands~1 and 2 of the MODIS ones, while, for~bands 3 to 7, the~spatial resolutions is of 500 m/pixel. As~GEE lets the users define the spatial resolution of the data to be acquired, each~area was covered exactly by one pixel in LANDSAT. In~MODIS, bands 3--7 contain information of the surrounding region of each one of the areas.}
For each area, we downloaded a time series, corresponding to the period covered from January, 2000 to August, 2016. In~order to fill gaps (principally due to the Landsat 7 failure in 2003), the~data were composite into monthly time series and is subsequently linearly~interpolated. 

\textls[-15]{Table~\ref{tab:bands-specification} presents information about the bands downloaded for both sensors, Landsat (\mbox{\url{https://www.usgs.gov/faqs/what-are-best-landsat-spectral-bands-use-my-research}} ({As of  3rd March 2020}))} and MODIS, associating the channel name with the band code and specifying the wavelength range covered by the band. We selected the channels blue, green, red, near~infrared (NIR) 1 and 2, and~short-wave infrared (SWIR) 1 and 2. Landsat TM products also normally provide information for the thermal infrared spectral region (LWIR), corresponding to band number 6. However, these specific SR products do not include this~information.

\begin{table}[H]
\centering
\caption[Band specification for Landsat and MODIS]{Specification of spectral bands for sensors Landsat and MODIS.}
\renewcommand{\arraystretch}{1.2}
\begin{tabular}{lccccl}
\toprule
\multicolumn{1}{c}{} & \multicolumn{2}{c}{\textbf{Landsat}} & \multicolumn{2}{c}{\textbf{MODIS}} \\ \midrule
\multicolumn{1}{c}{\textbf{Name}}    &  & \boldmath{$\lambda$ ($\upmu$}\textbf{m)} &  & \boldmath{$\lambda$ ($\upmu$}\textbf{m)} \\ \midrule
\multicolumn{1}{l}{\textbf{{Blue}}} & B1 & 0.45--0.52 & B3  & 0.46--0.48\\  
\multicolumn{1}{l}{\textbf{Green}} & B2 & 0.52--0.60 & B4  & 0.55--0.57\\
\multicolumn{1}{l}{\textbf{Red}} & B3 & 0.63--0.69 & B1 & 0.62--0.67\\
\multicolumn{1}{l}{\textbf{NIR}} & B4 & 0.76--0.90 & B2 & 0.84--0.88\\
\multicolumn{1}{l}{\textbf{NIR 2}} &-& - & B5 & 1.23--1.25\\
\multicolumn{1}{l}{\textbf{SWIR}} & B5 & 1.55--1.75 & B6 & 1.63--1.65\\
\multicolumn{1}{l}{\textbf{SWIR 2}} & B7 & 2.08--2.35 & B7 & 2.11--2.16\\
\bottomrule
\end{tabular}

\label{tab:bands-specification}
\end{table}
\unskip

\subsubsection{Baselines}
\label{subsubsec:traditional-indices}

The indices learned through the proposed method (GPVI) are compared to three of the most widely used vegetation indices to characterize forests and savannas~\cite{vina2011comparison}. Hereafter we will refer to these three methods as our baselines:

{Normalized difference vegetation index (NDVI)
	:} normalized difference between the red and near infrared channels to measure vegetation greenness~\cite{1974NASSP.351..309R}.
\begin{equation}
    \label{eq:ndvi}
NDVI = \displaystyle\frac{NIR - Red}{NIR + Red}
\end{equation}
    
{Enhanced vegetation index (EVI)
	:} enhancement of sensitivity adding the blue channel~\cite{huete1999modis}. Proposed to improve saturation problems occurring in NDVI, when representing forest heterogeneity.
\begin{equation}
        EVI = \displaystyle\frac{2.5 \times (NIR - Red)}{NIR + 6.0 \times Red - 7.5 \times Blue + 1}
    \end{equation}
    
{Two-band enhanced vegetation index (EVI2)
	:} calibration of EVI so it must not depend on the blue band, particularly because the blue band is not included in products of old sensors and normally presents noise problems~\cite{Jiang20083833}.
\begin{equation}
        EVI2 = \displaystyle\frac{2.5 \times (NIR - Red)}{NIR + 2.4 \times Red + 1}
    \end{equation}

\subsubsection{GP~Configuration}
\label{subsubsec:gp-conf}

Table~\ref{tab:gp_setup} shows a summary of the configuration of the GP algorithm. The~parameters of the formulas (i.e., leaves of the trees) are random real numbers between $0$ and $10^3$, and~the variables are spectral bands. The~possible operators of the formula (i.e., internal nodes) are addition ($+$), subtraction~($-$), multiplication (*), protected division ($\%$), protected square root ($srt()$), and~protected natural logarithm ($rlog()$), as~suggested by Koza~\cite{Koza:1992:GPP:138936} in order to avoid division by zero, imaginary roots, and~logarithm of negative~numbers. 

\begin{table}[H]
\centering
\caption{Genetic Programming (GP) parameters setup. The~three last operators mentioned stands for protected division ($\%$), protected square root ($srt()$) and protected natural logarithm ($rlog()$).}
\renewcommand{\arraystretch}{1.2}
\begin{tabular}{lc}
\toprule
 \multicolumn{1}{c}{\textbf{Parameter}} & \bfseries Value\\
\midrule
Population size & 100\\
Generations & 200\\
Operators (intern nodes) & \{+, -, *, \%, srt(), rlog()\}\\
Parameters (laves) & $\{b_i : 0 \le i < n\} \cup \{c_j \in [0, 10^{3}] \}$\\
Maximum initial tree depth & 6\\
Selection method & Tournament $\times\ 3$\\
Crossover rate & $0.9$\\
Mutation rate & $0.1$\\
\bottomrule
\end{tabular}

\label{tab:gp_setup}
\end{table}

The experiments were performed with $100$ individuals and $200$ generations. New randomly generated trees will not have a depth greater than 6. The~selection method is tournament with three individuals. It consists of randomly selecting three individuals from the population and allowing the best one to go to crossover, as~many times as a new population is completed. Once two individuals are chosen, they have a probability of $0.9$ to cross their genetic information and create new individuals. Every new individual in the population has a probability of $0.1$ to mutate. The~algorithm must stop when $200$ generations are reached. The~learning process for all classification problems using this configuration converged correctly, so more complex configurations, e.g.,~more iterations or a higher number of generations were not likely to improve~performance.

\subsubsection{Evaluation~Protocol}
\label{subsubsec:eval-protocol-tropical}

GPVI is optimized to discriminate raw pixels, rather than whole time series. We considered the pixels only corresponding to the first five years (from 2000 to 2004) of the collected time series as training set for the GP framework, without~preserving their temporal relation, since GPVI is intended to be time-independent. Thus, each pixel in each location (281 in total) and time stamp (12 months $\times$ five years) is a sample. Once obtained, the~GPVI is used to classify the rest of the data, used as test set. The~classification performance on the test set (from 2005 to 2016) is reported in the following~section.

\subsubsection{Results}

The performance of GPVI and the indices used as baselines are summarized in Table~\ref{tab:acc-f-s}. The~columns marked with {\bf Prod.}, {\bf User}, and~{\bf Acc.} correspond to the producer's, user's, and~normalized accuracies, respectively. The~producer's accuracy for each class is the rate of correctly classified samples that belong to that class, while the user's accuracy is the rate of correctly classified samples that are classified by the system as such. The~normalized accuracy is the average of the producer's accuracy of all the classes, which is a total accuracy (rate of correctly classified samples) giving all of the classes the same weight, regardless of the number of samples that belong to each one. The~results obtained for Landsat and MODIS sensors can be~contrasted.

\begin{table}[H]
\centering
\caption{Normalized accuracy (\textbf{Acc.}) and user's (\textbf{User}) and producer's \textbf{(Prod.) accuracy} per class for forest/savanna~discrimination.}
\renewcommand{\arraystretch}{1.2}
\begin{tabular}{ccccccccccc}
\toprule
\multicolumn{1}{c}{} & \multicolumn{5}{c}{\textbf{Landsat}} & \multicolumn{5}{c}{\textbf{MODIS}} \\ \midrule
\multicolumn{1}{c}{} & \multicolumn{2}{c}{\textbf{Forest}} & \multicolumn{2}{c}{\textbf{Savanna}} & \multicolumn{1}{c}{} & \multicolumn{2}{c}{\textbf{Forest}} & \multicolumn{2}{c}{\textbf{Savanna}} & \multicolumn{1}{c}{} \\ \midrule
\multicolumn{1}{c}{} & \textbf{Prod.} & \textbf{User} & \textbf{Prod.} & \textbf{User} & \textbf{Acc.} & \textbf{Prod.} & \textbf{User} & \textbf{Prod.} & \textbf{User} & \textbf{Acc.} \\ \midrule
\multicolumn{1}{l}{NDVI} &  89.28 & 86.60 & 90.49 & 92.44 & 89.89  & 78.71 & 80.19 & 86.44 & 85.37 & 82.58  \\
\multicolumn{1}{l}{EVI} & 87.75 & 85.64 & 89.88 & 91.42 & 88.81 &  80.90 & 83.79 & 89.10 & 86.98 & 85.00  \\
\multicolumn{1}{l}{EVI2} &  87.34 & 85.14 & 89.48 & 91.10 & 88.41 & 78.24 & 81.90 & 87.93 & 85.27 & 83.09  \\ \midrule
\multicolumn{1}{l}{GPVI} &  96.38 & 93.59 & 95.46 & 97.46 & 95.92 & 88.30 & 86.16 & 91.09 & 91.08 & 89.69 \\ \bottomrule
\end{tabular}
\label{tab:acc-f-s}
\end{table}

The baselines presented similar performances and were clearly outperformed by GPVI ($6\%$ better as compared to Landsat and $4\%$ to MODIS). In~general, the~performance for Landsat was better than the one for MODIS, and~the rate of correctly classified savannas was higher than the rate of correctly classified forests. Regarding user's accuracy, a~clear superiority of GPVI over the baselines was~observed.

{It is important to remark that a lower performance of GPVI and the baselines is expected for MODIS, as~bands 3 to 7 have lower spatial resolution, which forced the data collection process to include information of the surrounding of the areas, which can be considered as noise.}

\begin{figure}[H]
\centering
\begin{subfigure}{0.49\textwidth}
\centering \footnotesize
NDVI
\vspace*{-0.5\baselineskip}

\includegraphics[width=1\columnwidth]{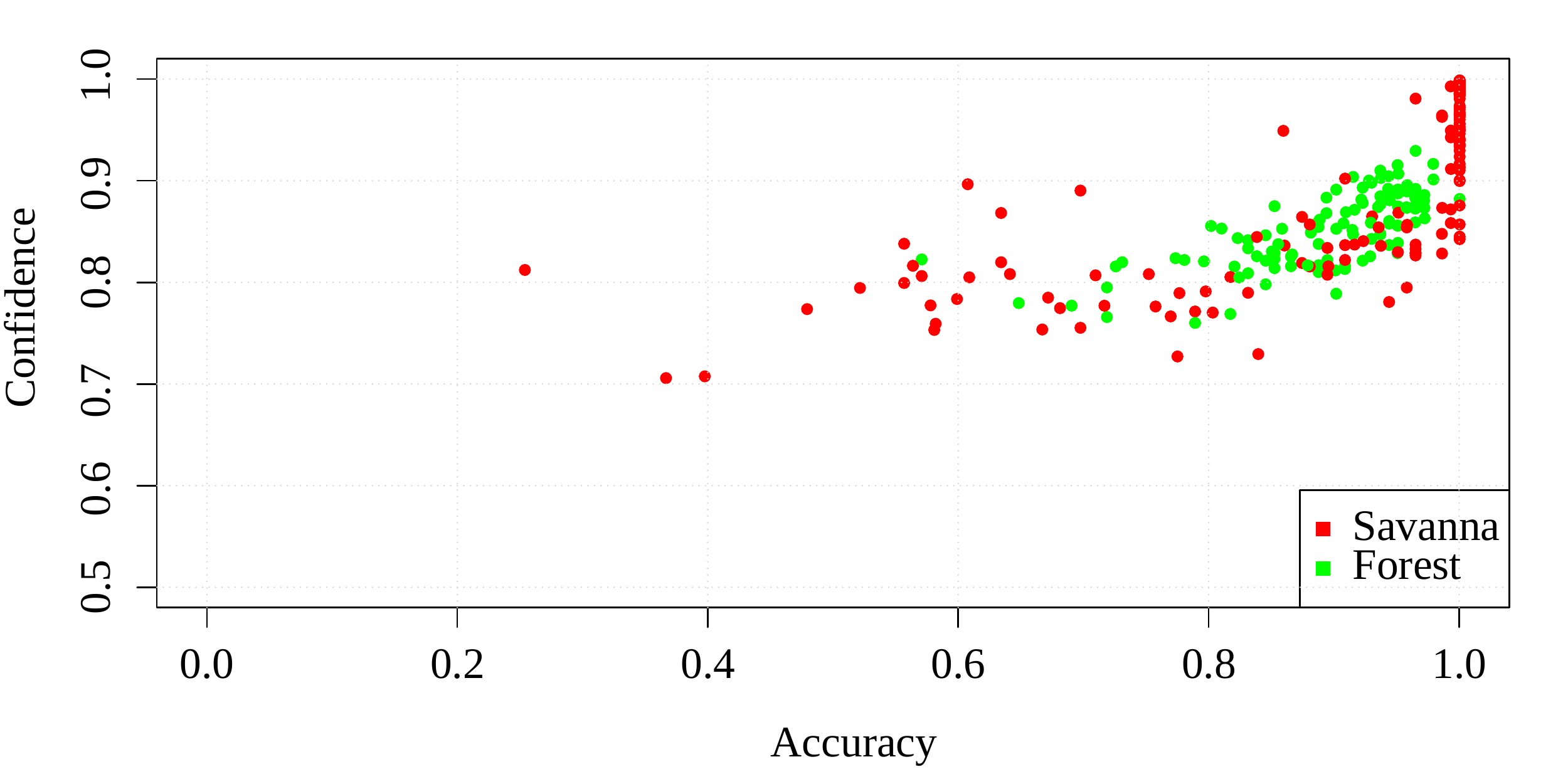}
\end{subfigure}
\begin{subfigure}{0.49\textwidth}
\centering \footnotesize
EVI
\vspace*{-0.5\baselineskip}

\includegraphics[width=1\columnwidth]{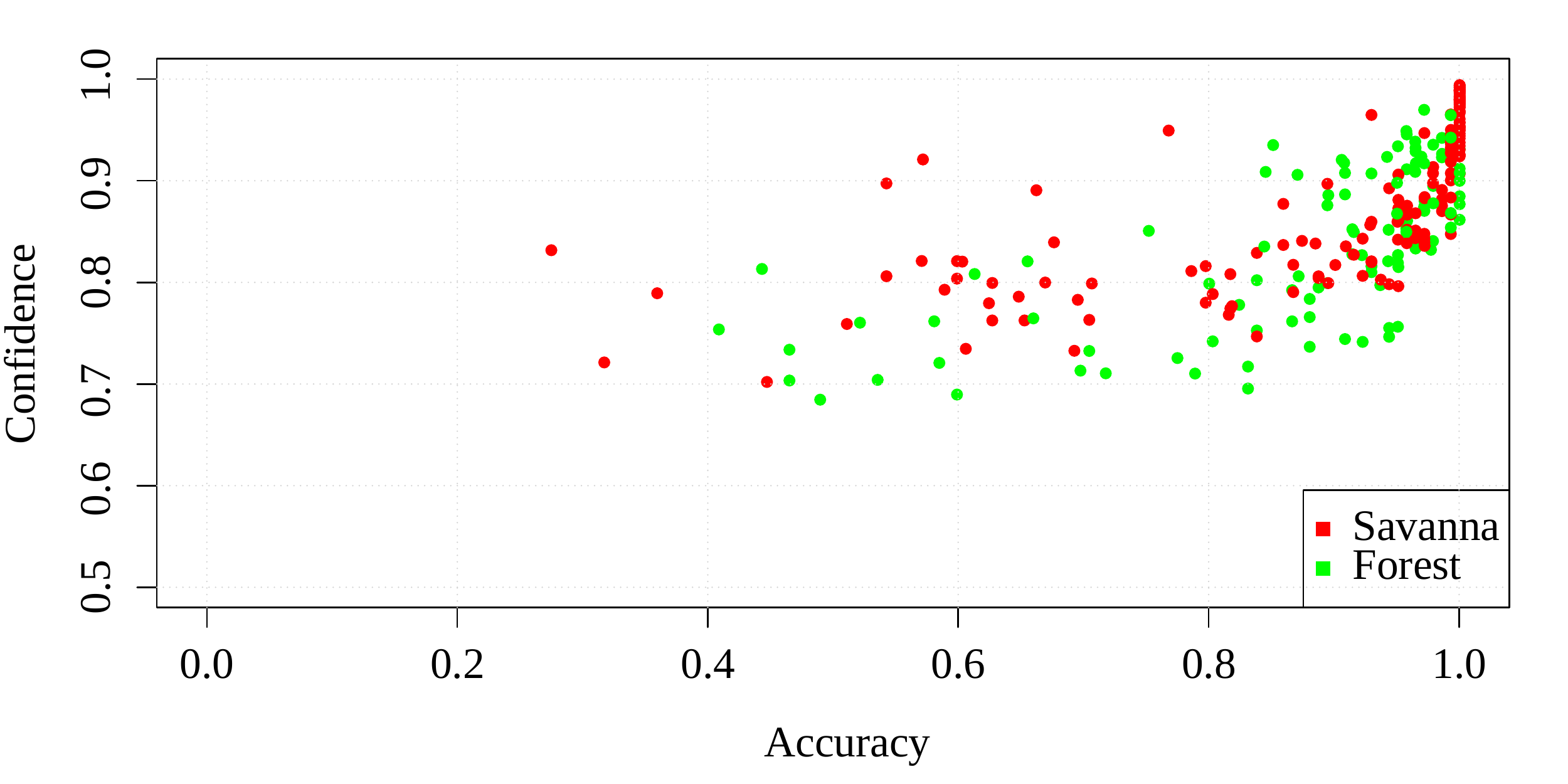}
\end{subfigure}

\vspace*{0.5\baselineskip}

\begin{subfigure}{0.49\textwidth}
\centering \footnotesize
EVI2
\vspace*{-0.5\baselineskip}

\includegraphics[width=1\columnwidth]{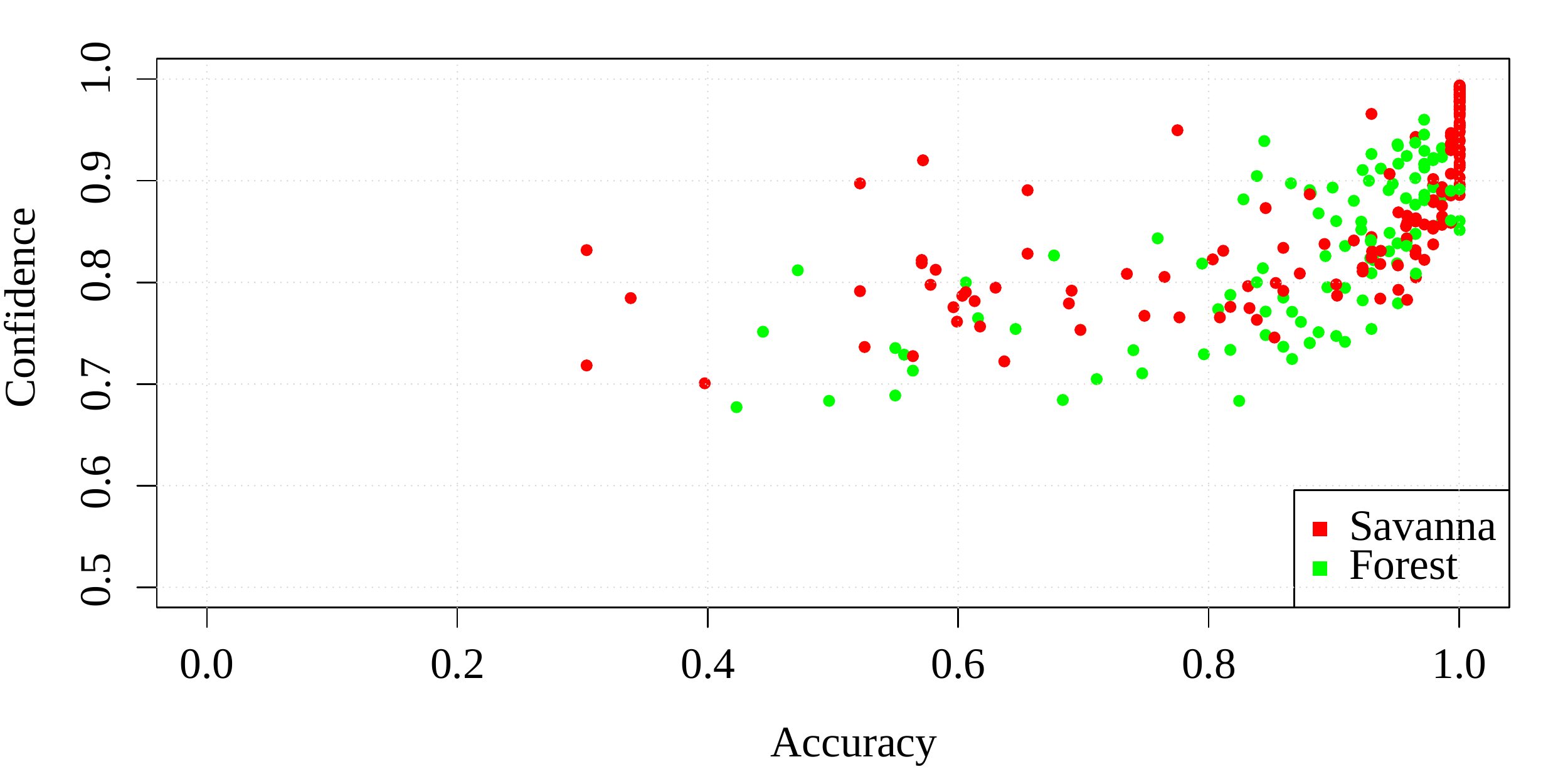}
\end{subfigure}
\begin{subfigure}{0.49\textwidth}
\centering \footnotesize
GPVI
\vspace*{-0.5\baselineskip}

\includegraphics[width=1\columnwidth]{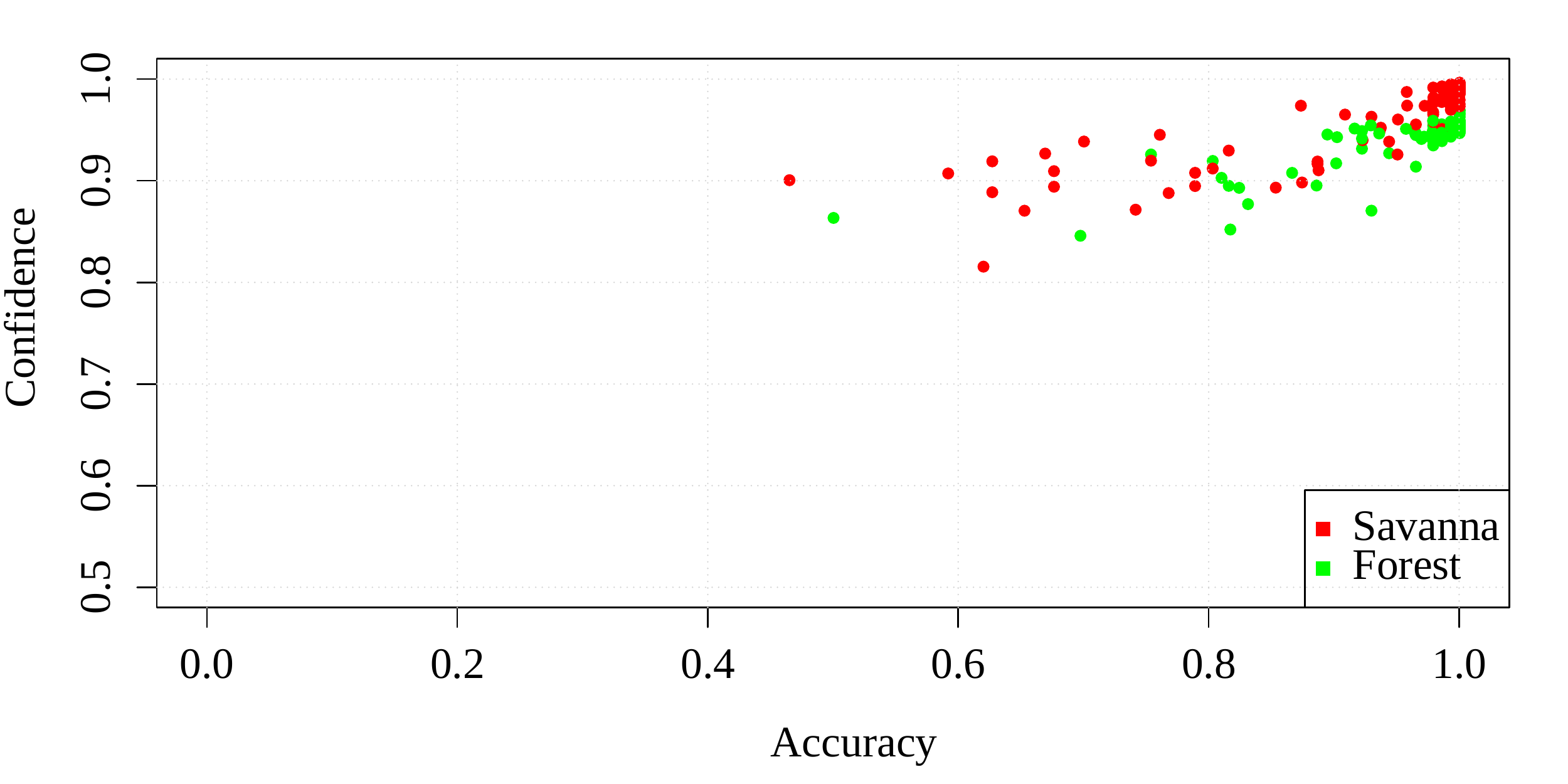}
\end{subfigure}
\caption{Confidence vs. accuracy in each area (Landsat).}
\label{fig:conf-cor-landsat}
\end{figure}

\begin{figure}[H]
\centering
\begin{subfigure}{0.49\textwidth}
\centering \footnotesize
NDVI
\vspace*{-0.5\baselineskip}

\includegraphics[width=1\columnwidth]{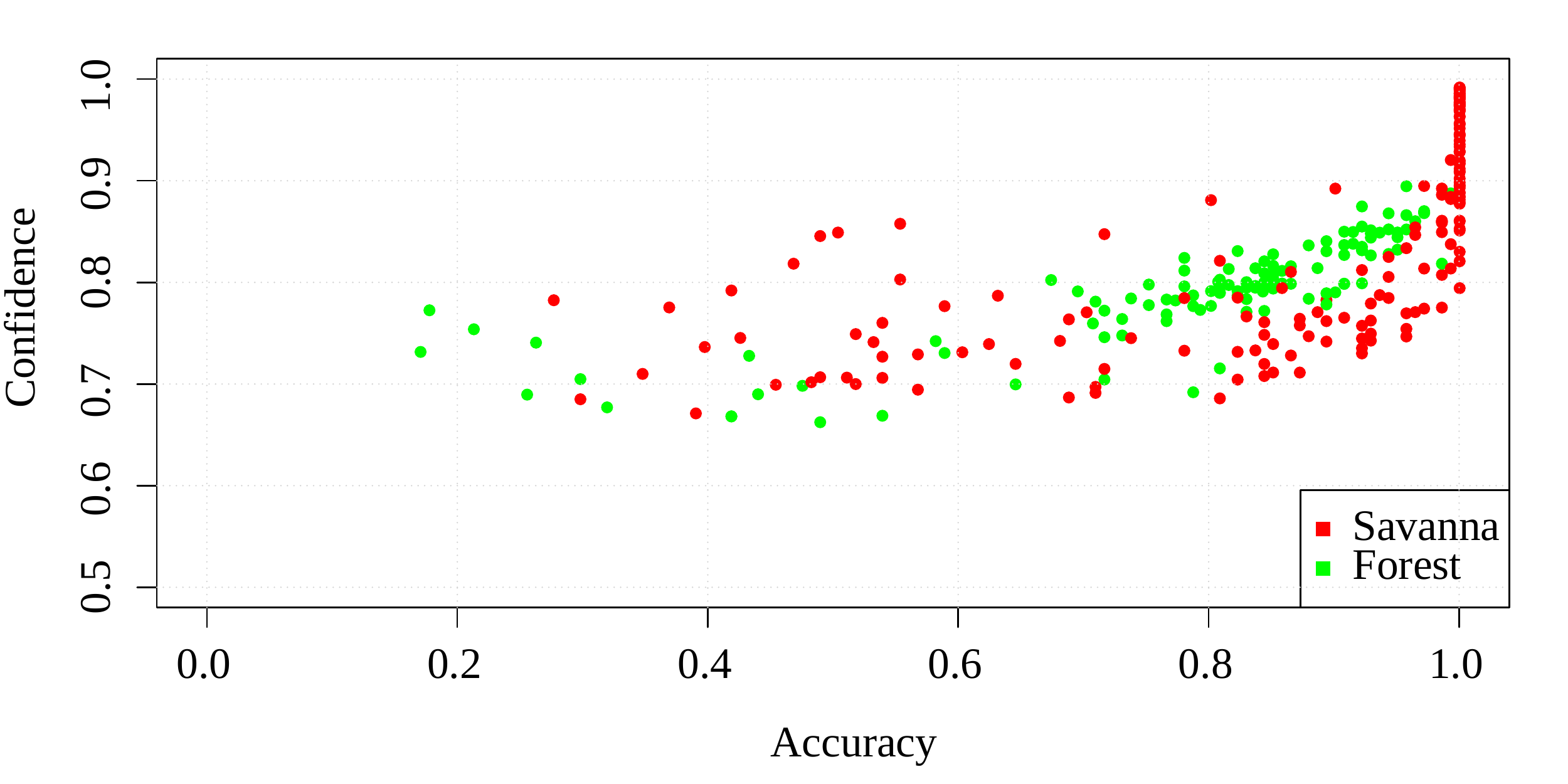}
\end{subfigure}
\begin{subfigure}{0.49\textwidth}
\centering \footnotesize
EVI
\vspace*{-0.5\baselineskip}

\vspace*{0.5\baselineskip}

\includegraphics[width=1\columnwidth]{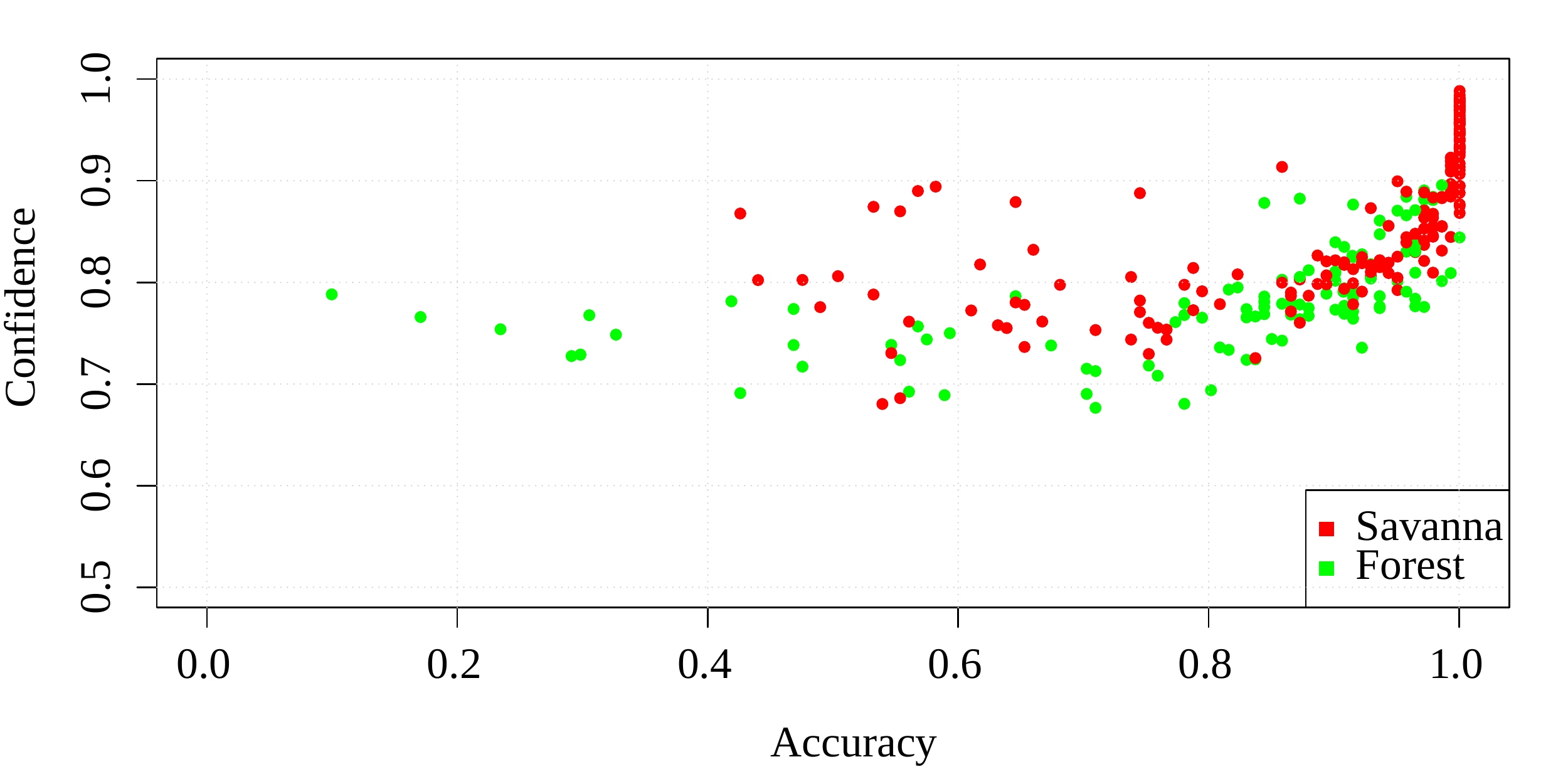}
\end{subfigure}
\begin{subfigure}{0.49\textwidth}
\centering \footnotesize
EVI2
\vspace*{-0.5\baselineskip}

\includegraphics[width=1\columnwidth]{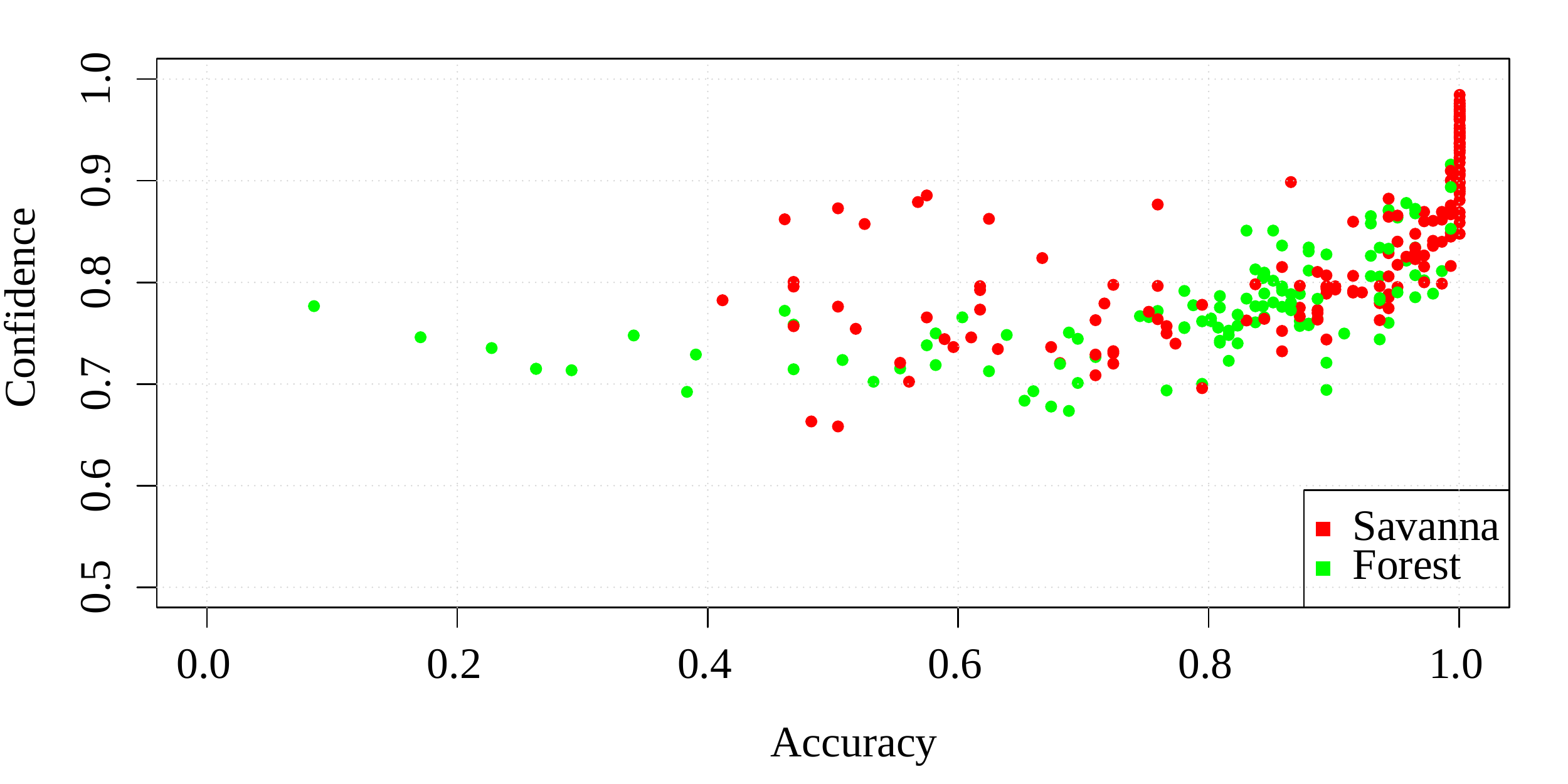}
\end{subfigure}
\begin{subfigure}{0.49\textwidth}
\centering \footnotesize
GPVI
\vspace*{-0.5\baselineskip}

\includegraphics[width=1\columnwidth]{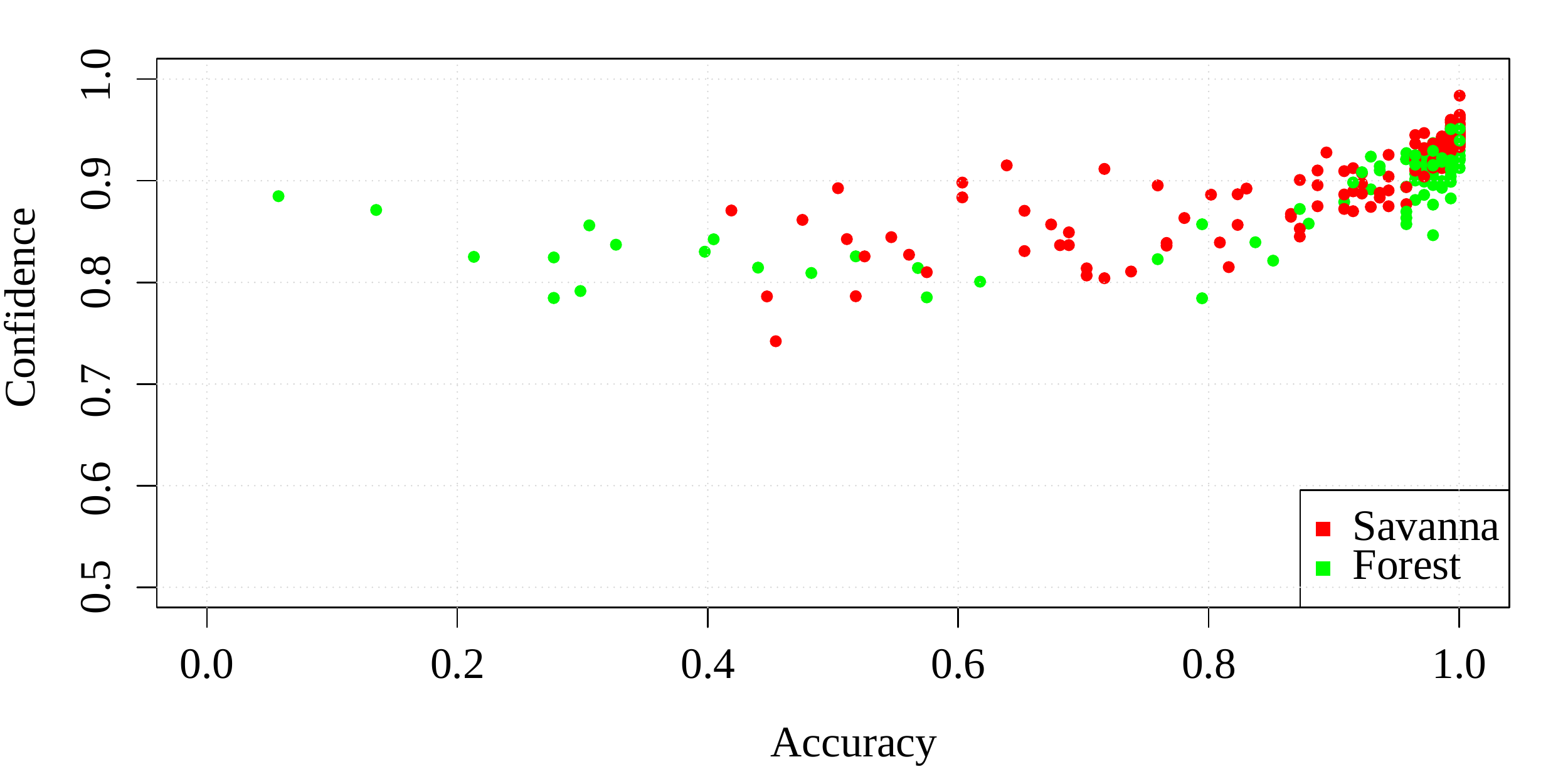}
\end{subfigure}
\caption{Confidence vs. accuracy in each area (MODIS).}
\label{fig:conf-cor-modis}
\end{figure}

An additional experiment tests the confidence of the classification of the spectral indices. For~this, we took the one-band pixel values, yielded by each spectral index, for~all of the locations, {trained~a Logistic Regression classification algorithm with them~\cite{hastie01statisticallearning} ([Ch.~4]), and~used its predicted class probability given a data point as a confidence score}, since the class that is assigned to the areas depends on whether this value is under or over $0.5$. Thus, scores that are close to this value are more likely to be classified incorrectly, since the algorithm is totally uncertain of the class that it assigns to the sample. Figures~\ref{fig:conf-cor-landsat}~and~\ref{fig:conf-cor-modis} show scatterplots of mean accuracy vs. mean confidence score obtained in each area, for~Landsat and MODIS, respectively. It can be seen that GPVI presented, on~average, higher~confidence values than the baselines, which means less confusion at classification time.
{The~fact that savannas and forest can transit into each other's states~\cite{Dantas2016}, can explain those areas that, in~general, obtained low confidence and low accuracy, since a transitional location presents features of both biomes, leading to a higher confusion in the classification algorithm.}
Those areas presenting low accuracy and high confidence in various experiments could also result from mislabeled~samples. 

\unskip

\unskip

\subsection{Gpvi for Finer-Scale Vegetation Type~Classification}
\label{subsec:vegetation-type}

To address~\ref{rq2}, we consider two separate classification problems that are harder than forest/savanna discrimination, as~they deal with the discrimination of vegetation types (subcategories~of savannas and forests with respect to their physiognomy): evergreen vs. semi-deciduous forests, and~typical vs. forested savannas. In~this setting, traditional indices often saturate and, normally, are not even considered in analyses, because~they lack representation of within-biome differences \mbox{(i.e., they consider} the whole biome as an homogeneous landscape)~\cite{Xue2017JournalofSensors,vina2011comparison,SCHULTZ2016318}.

The data used for this set of experiments, as~well as the baselines, the~parameters used in the GP framework, and~the evaluation protocol, are exactly the same as the ones that are used for the forest/savanna classification problem (Section~\ref{subsec:biome}).

The performance of GPVI and the indices used as baselines are summarized in Tables~\ref{tab:acc-f} and~\ref{tab:acc-s}. The~results obtained for Landsat and MODIS sensors can be~contrasted.

\begin{table}[H]
\centering
\caption{Normalized accuracies (\textbf{Acc.}) and user's (\textbf{User}), and~producer's \textbf{(Prod.)} accuracies per class for evergreen forest (\textbf{EGF})/semi-deciduous forest (\textbf{SDF}) discrimination.}
\renewcommand{\arraystretch}{1.2}
\begin{tabular}{ccccccccccc}
\toprule
\multicolumn{1}{c}{} & \multicolumn{5}{c}{\textbf{Landsat}} & \multicolumn{5}{c}{\textbf{MODIS}} \\ \midrule
\multicolumn{1}{c}{} & \multicolumn{2}{c}{\textbf{EGF}} & \multicolumn{2}{c}{\textbf{SDF}} & \multicolumn{1}{c}{} & \multicolumn{2}{c}{\textbf{EGF}} & \multicolumn{2}{c}{\textbf{SDF}} & \multicolumn{1}{c}{} \\ \midrule
\multicolumn{1}{c}{} & \textbf{Prod.} & \textbf{User} & \textbf{Prod.} & \textbf{User} & \textbf{Acc.} & \textbf{Prod.} & \textbf{User} & \textbf{Prod.} & \textbf{User} & \textbf{Acc.} \\ \midrule
\multicolumn{1}{l}{NDVI} &  67.04 & 92.94 & 39.95 & 9.18 & 53.50 & 75.17 & 75.00 & 61.61 & 12.95 & 68.39 \\
\multicolumn{1}{l}{EVI} & 64.50 & 95.21 & 61.38 & 12.72 & 62.94 & 72.19 & 96.73 & 63.39 & 13.02 & 67.79 \\
\multicolumn{1}{l}{EVI2} & 63.93 & 94.55 & 56.61 & 11.61 & 60.27 & 69.95 & 96.41 & 61.08 & 12.20 & 65.52 \\ \midrule
\multicolumn{1}{l}{GPVI} & 78.40 & 96.03 & 61.90 & 19.34 & 70.15 & 84.44 & 98.51 & 78.72 & 23.21 & 81.58 \\ \bottomrule
\end{tabular}
\label{tab:acc-f}
\end{table}
\unskip

\begin{table}[H]
\caption{Normalized accuracies (\textbf{Acc.}) and user's (\textbf{User}), and~producer's \textbf{(Prod.)} accuracies per class for typical savanna (\textbf{TS})/forested savanna (\textbf{FS}) discrimination.}
\centering
\renewcommand{\arraystretch}{1.2}
\begin{tabular}{ccccccccccc}
\toprule
\multicolumn{1}{c}{} & \multicolumn{5}{c}{\textbf{Landsat}} & \multicolumn{5}{c}{\textbf{MODIS}} \\ \midrule
\multicolumn{1}{c}{} & \multicolumn{2}{c}{\textbf{TS}} & \multicolumn{2}{c}{\textbf{FS}} & \multicolumn{1}{c}{} & \multicolumn{2}{c}{\textbf{TS}} & \multicolumn{2}{c}{\textbf{FS}} & \multicolumn{1}{c}{} \\ \midrule
\multicolumn{1}{c}{} & \textbf{Prod.} & \textbf{User} & \textbf{Prod.} & \textbf{User} & \textbf{Acc.} & \textbf{Prod.} & \textbf{User} & \textbf{Prod.} & \textbf{User} & \textbf{Acc.} \\ \midrule
\multicolumn{1}{l}{NDVI} & 57.92 & 96.96 & 73.70 & 10.60 & 65.81 & 59.89 & 96.80 & 70.99 & 10.88 & 65.44 \\
\multicolumn{1}{l}{EVI} & 58.37 & 96.17 & 65.61 & 9.72 & 61.99 & 62.91 & 96.03 & 62.28 & 10.33 & 62.60 \\
\multicolumn{1}{l}{EVI2} & 58.36 & 96.17 & 66.17 & 9.71 & 62.27 & 62.68 & 96.01 & 62.48 & 10.26 & 62.58 \\ \midrule
\multicolumn{1}{l}{GPVI} & 71.06 & 97.66 & 74.89 & 14.98 & 72.98 & 58.62 & 97.32 & 76.34 & 11.28 & 67.48 \\ \bottomrule
\end{tabular}
\label{tab:acc-s}
\end{table}

For the evergreen forest/semi-deciduous forest discrimination (Table~\ref{tab:acc-f}), the~baselines presented similar performances when compared to GPVI, except~for NDVI from Landsat, which presented a much inferior performance than the other baselines. In~general, the~baselines behaved similarly for both sensors, and~they were outperformed by GPVI by about $7\%$ for Landsat, and~by approximately $13\%$ for MODIS. The~performance of GPVI using MODIS data was much higher than Landsat, and~evergreen forest was classified with a higher rate of correctly classified areas. The~user's accuracy shows an evident bias towards the class evergreen forest for all of the indices. However, this bias is slightly lower for~GPVI.

For the typical savanna/forested savanna discrimination (Table~\ref{tab:acc-s}), the~baselines presented similar performances for both sensors, and~they were outperformed by GPVI by about $7\%$ for Landsat, and~by about $2\%$ for MODIS. The~performance of GPVI using Landsat data was superior than using MODIS, and~forested savanna obtained a higher rate of correctly classified areas only for Landsat; for MODIS, both classes obtained similar rates. The~user's accuracy shows an evident bias towards typical savannas for all of the indices. However, this bias is slightly lower for~GPVI.

\subsection{Gpvi for Time-Series Analysis And~Classification}
\label{subsec:time-series}

To address~\ref{rq3}, we apply GPVI, along with the baselines, to~the data collected, but~now maintaining its temporal relation. Therefore, we are now not working with raw pixels, but~with whole time series. For~this set of experiments, there was no index-learning (i.e., training) stage, since we took the same indices that were learned in the experiments for raw-pixel-based forest/savanna classification (Section~\ref{subsec:biome}), and~evergreen forest/semi-deciduous forest and typical savanna/forested savanna classification (Section~\ref{subsec:vegetation-type}). The~only difference now is that they will be applied to the multispectral time series in order to yield one-dimensional time~series.

The data used for this set of experiments, as~well as the baselines, are exactly the same as the ones used for the raw-pixelwise forest/savanna classification problem (Section~\ref{subsec:biome}).

\subsubsection{Experimental~Protocol}

We use a slightly different protocol from the raw-pixel classification experiments, due to the amount of data available. For~pixel classification, each pixel ($281$ in total) in each location at each time stamp ($12$ months $\times$ 5 years) is a sample, so there are enough data ($281 \times 12 \times 5 =$ 16,860) to provide a reliable comparison between GPVI and other indices. For~time series classification, on~the other hand, the~number of samples is substantially smaller, 25 areas had to be discarded due to a large amount of gaps, preventing a reliable linear interpolation to fill them. This yields $256$ areas, which means a sample of $256$ sites and, thus, a~cross validation protocol is~used.

We apply the GPVIs learned in the raw-pixel-classification phase on the time series of multispectral pixels to yield a one-dimensional time series per pixel. We do the same with the baseline indices. The~resulting series are used as an input of a time-series classification~scheme.

The time series classification set-up considers the entire time series of each area as a sample, and~splits the datasets to settle a $5\times 2$-fold cross validation protocol. In~this protocol, we randomly sample half of the series, preserving class distribution, and~use this half to train our classification algorithm; the other half is used for validation. The~subsets are then swapped (the one used for training was now used for validation and vice-versa). This process is repeated five times, which results in a total of ten classification~experiments.

We use the well-known 1-NN with Dynamic Time Wrapping (DTW) distance~\cite{Berndt1994} time-series classification algorithm to perform the experiments.
{We performed a series of statistical tests in order to compare GPVI with each one of the baselines, as~suggested by the literature on classification algorithms~\cite{Demvsar2006, Dietterich1998}.
First, we performed a Friedman test, a~non-parametric test across multiple measures, which, in~this case, are each one of the $5\times 2$-fold accuracies from GPVI and the three baselines.
We~complemented our analysis with Wilcoxon Signed Rank tests among every pair of methods. This~test can only be done between two sets of experiments, while assuming that each experiment in a population is paired to another experiment of the other population, and~it is also non-parametric.
For the Friedman and Wilcoxon tests, we considered a significance level of $0.05$.
Finally, to~not underestimate the $p$-values yielded by the Wilcoxon tests, we applied Bonferroni post hoc adjustment on the obtained $p$-values.}

Up to this point, we have shown the effectiveness of GPVI for pixelwise discrimination, without~considering temporal information. However, in~this section, we show how it behaves using time series, first with a visual comparison of GPVI with the other indices and then with the time-series classification set-up described~above.

\subsubsection{Time Series Analysis by Visual~Assessment}

The time series for GPVI and the baselines are graphically represented, taking the mean of the pixel values of all areas from the same class in each time stamp, so it is desirable that the curves corresponding to different classes are far away from each other. The~instantaneous standard deviation of each class are also shown around the corresponding curve, only for the forest-savanna discrimination, for~visualization purposes, since the other classification problems presented less separability. It is important to point that, since GPVI is learned with many degrees of freedom, the~values are not necessarily bounded like NDVI, for~example, comprising values between $-$1 and~1.

\begin{figure}[H]
\centering
\includegraphics[width=.49\columnwidth]{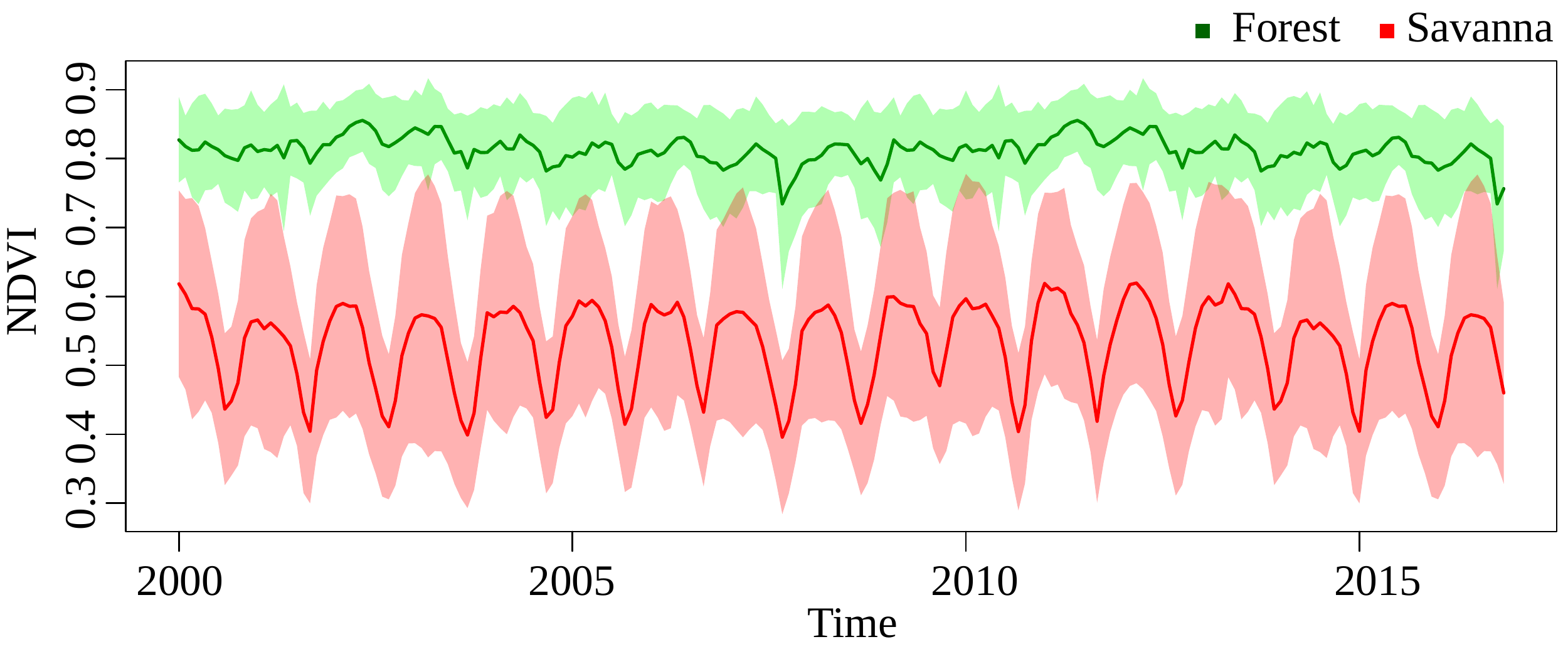}
\includegraphics[width=.49\columnwidth]{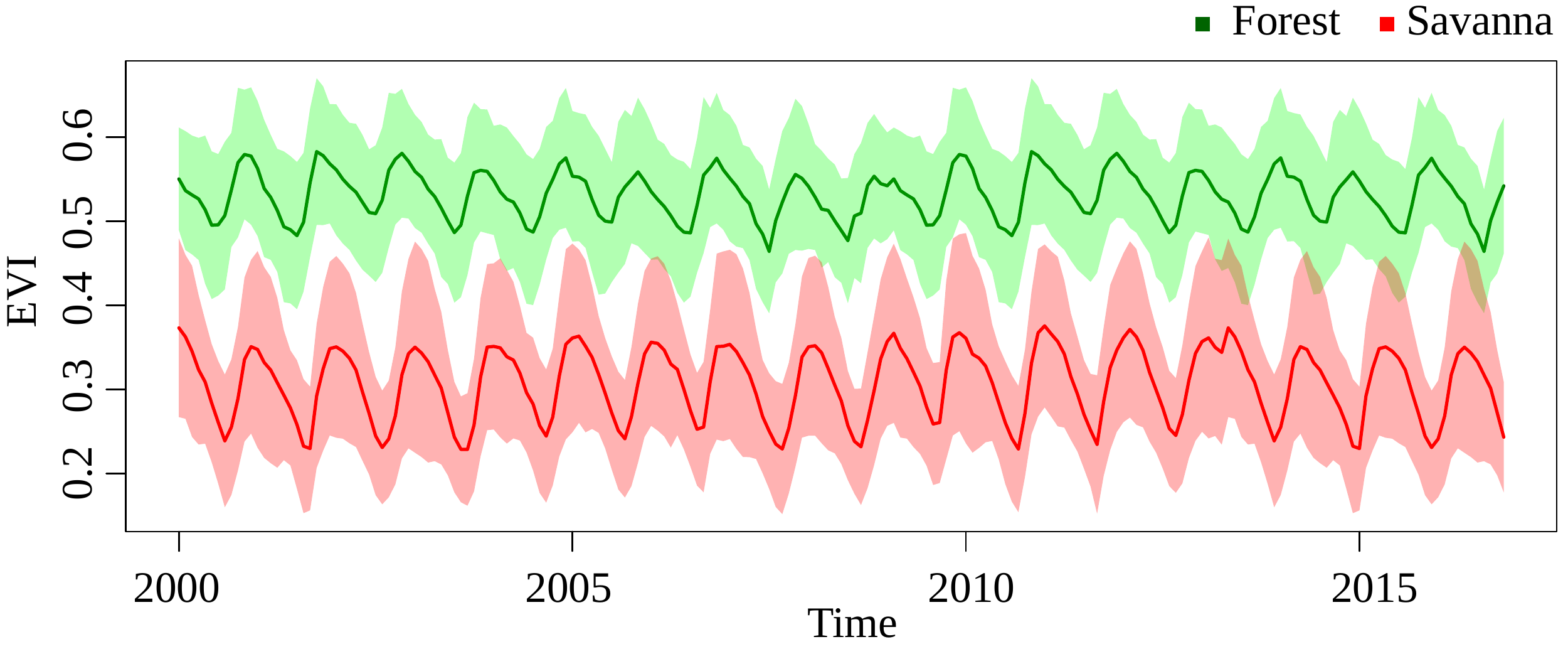}
\includegraphics[width=.49\columnwidth]{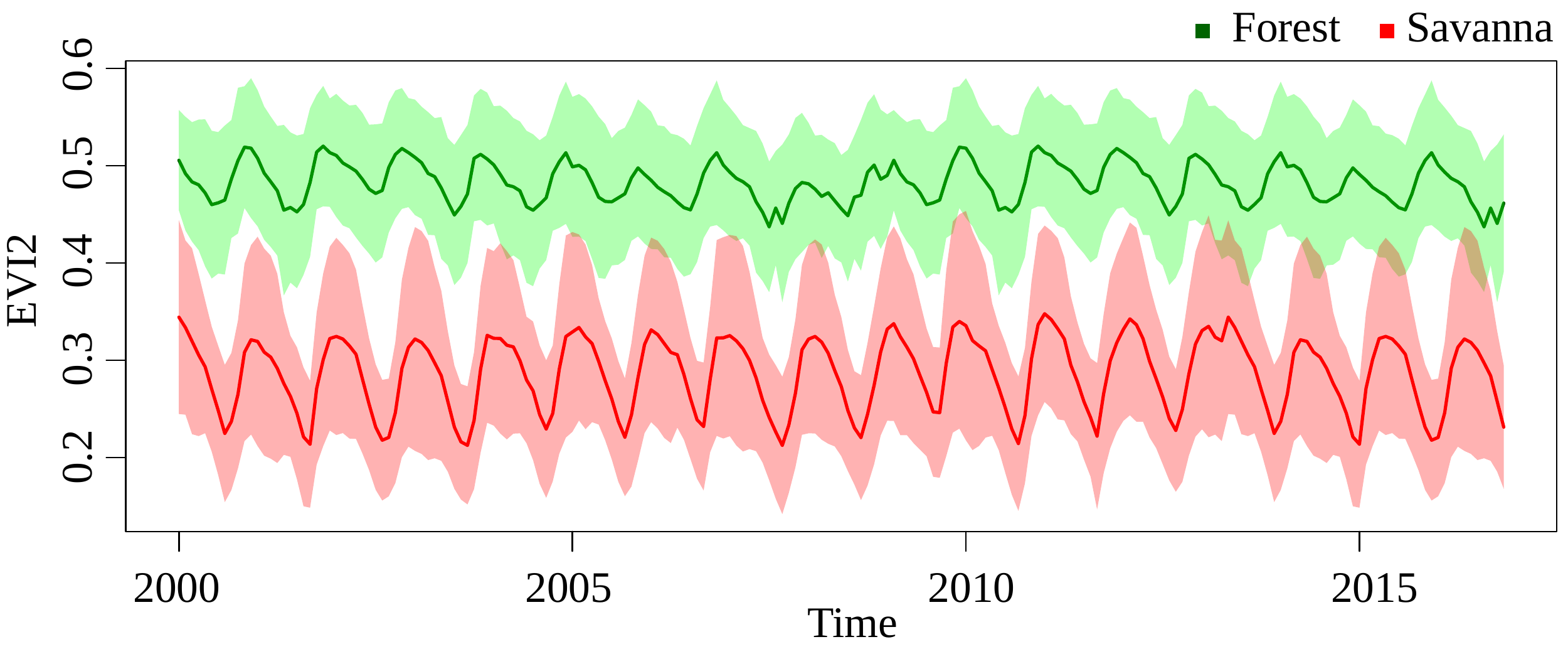}
\includegraphics[width=.49\columnwidth]{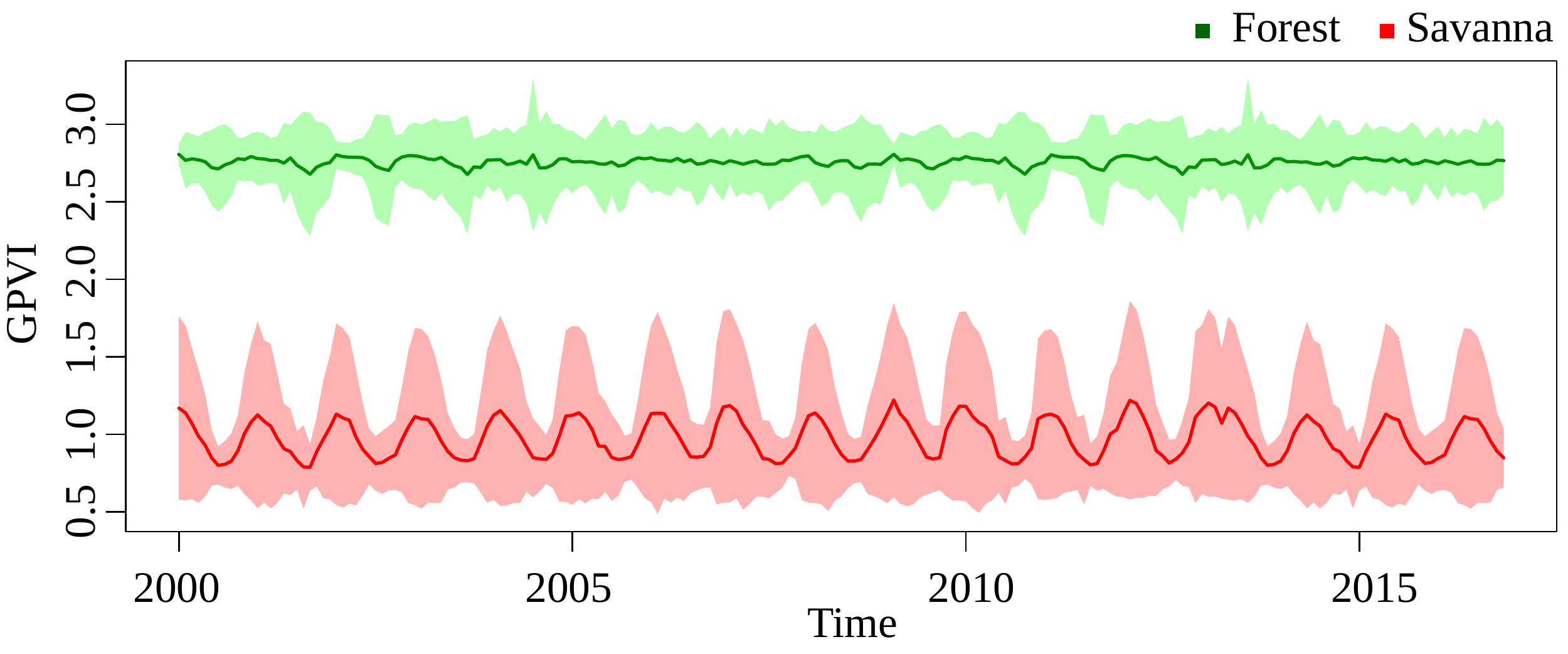}
\caption{Time series for forest/savanna discrimination (Landsat). {Shaded areas indicate the standard deviation of the index scores among all the data points for a given timestamp.}}
\label{fig:ts-landsat-f-s}
\end{figure}

\begin{figure}[H]
\centering
\includegraphics[width=.49\columnwidth]{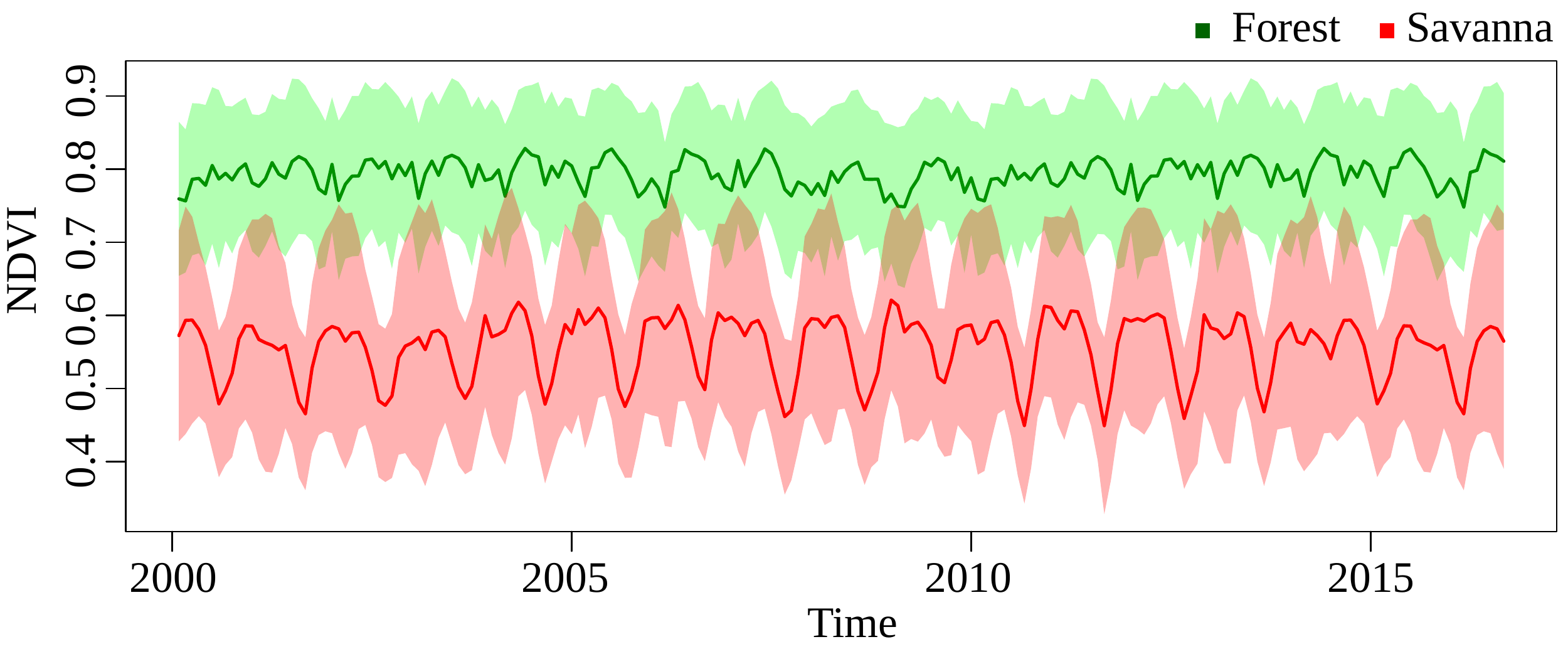}
\includegraphics[width=.49\columnwidth]{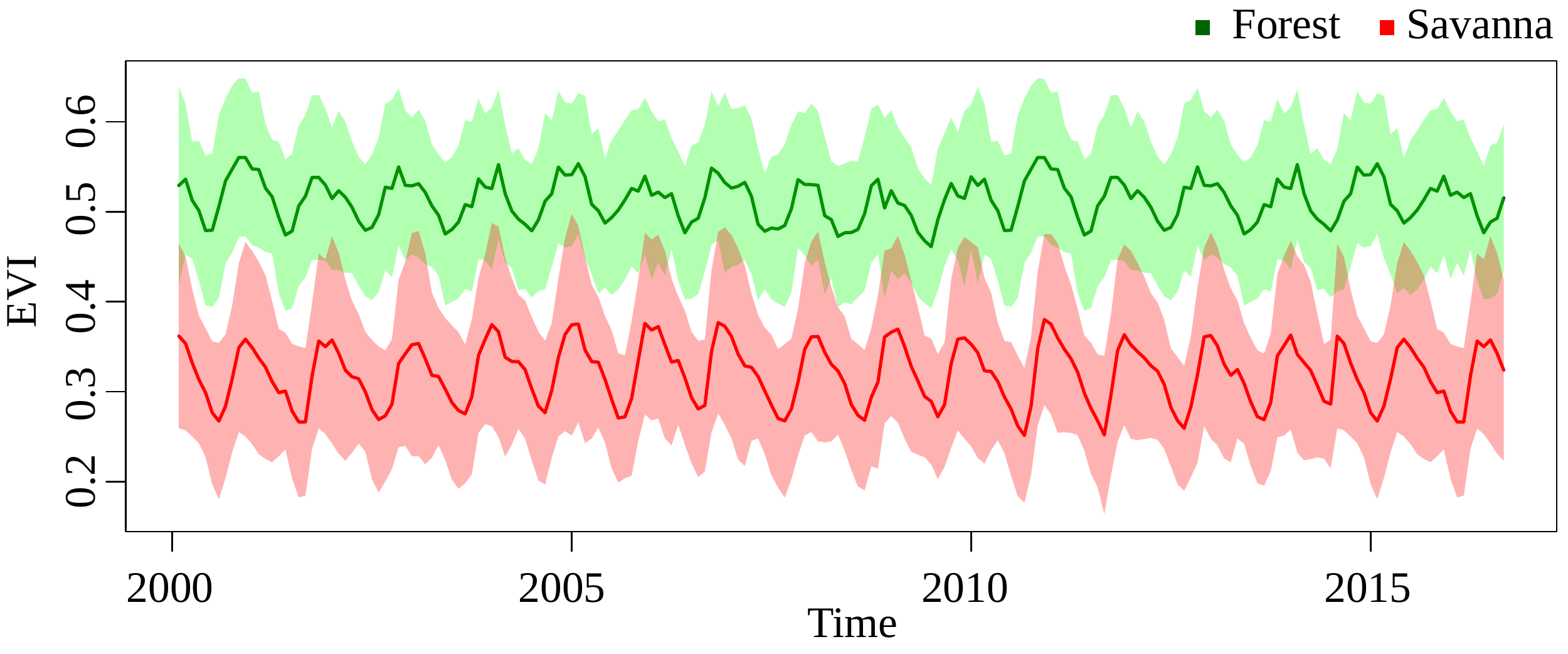}
\includegraphics[width=.49\columnwidth]{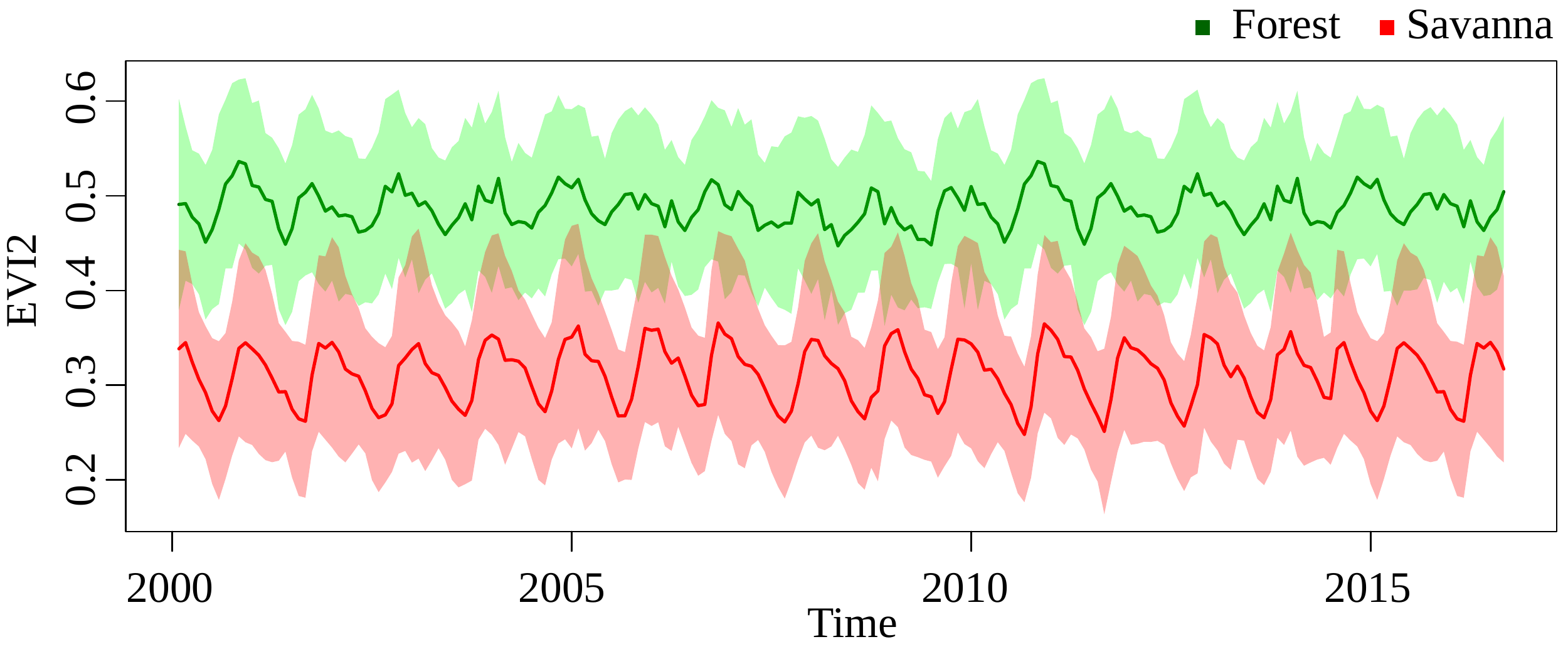}
\includegraphics[width=.49\columnwidth]{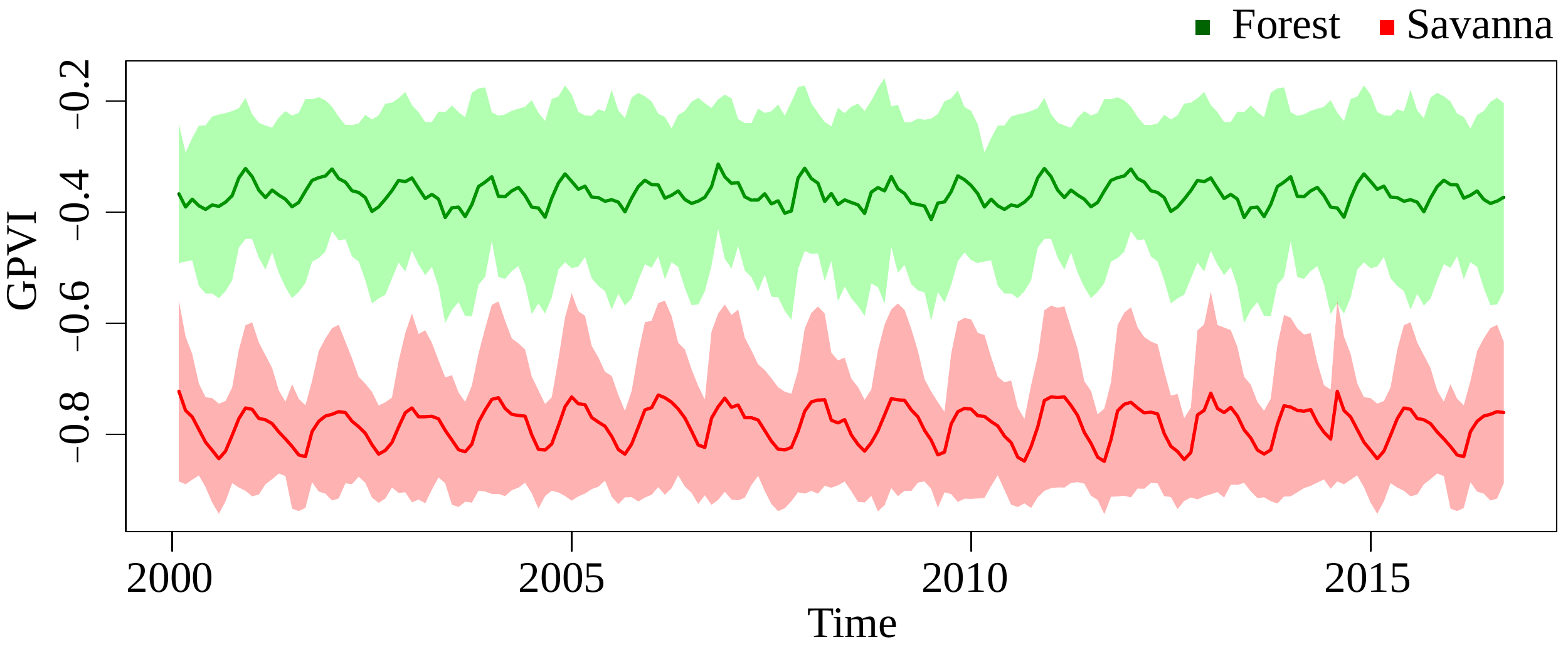}
\caption{Time series for forest/savanna discrimination (MODIS). {Shaded areas indicate the standard deviation of the index scores among all of the data points for a given timestamp.}}
\label{fig:ts-modis-f-s}
\end{figure}

Figures~\ref{fig:ts-landsat-f-s}~and~\ref{fig:ts-modis-f-s} show the time series for Landsat and MODIS, respectively, in~the forest/savanna classification problem. Landsat data present a greater separation between the series than MODIS, for~GPVI and the baselines. It is clear how GPVI makes it easier to discriminate the values of pixels for the two classes, regardless of the time~stamp.

\unskip

\begin{figure}[H]
\centering
\includegraphics[width=.49\columnwidth]{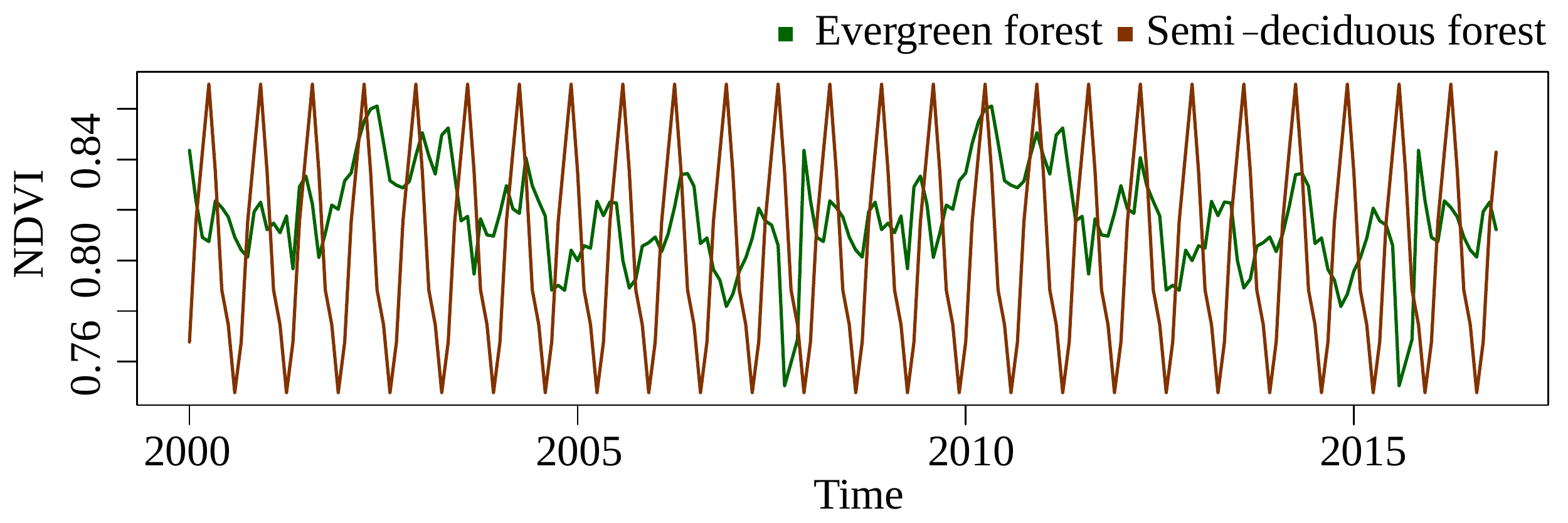}
\includegraphics[width=.49\columnwidth]{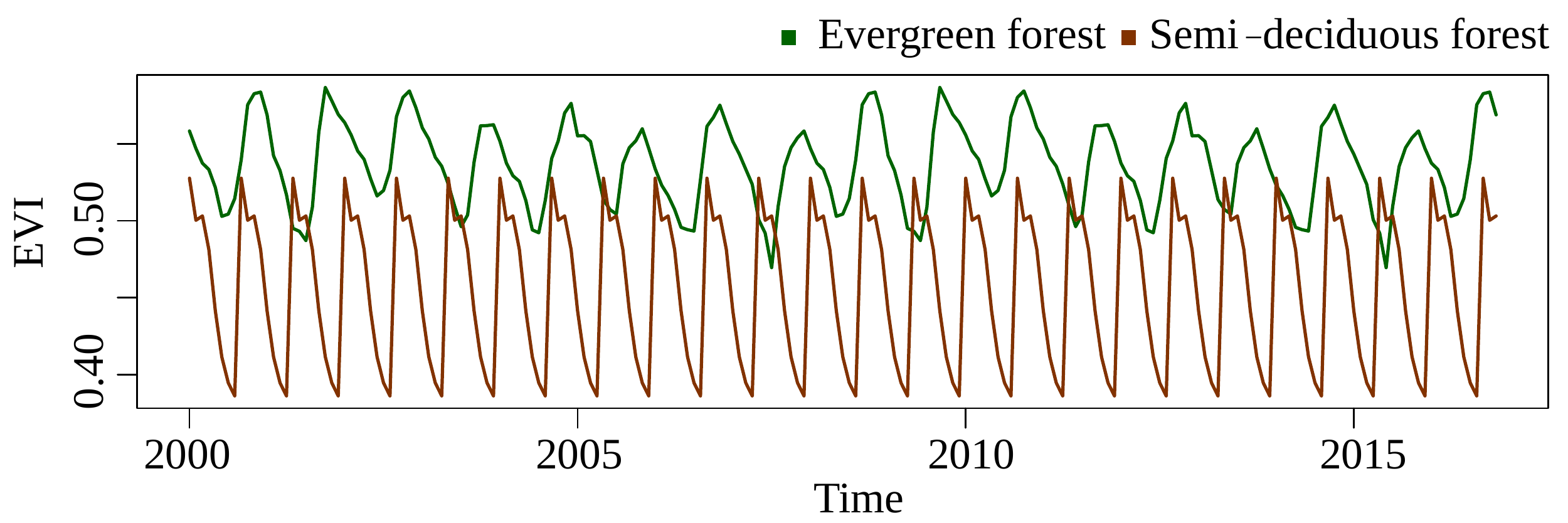}
\includegraphics[width=.49\columnwidth]{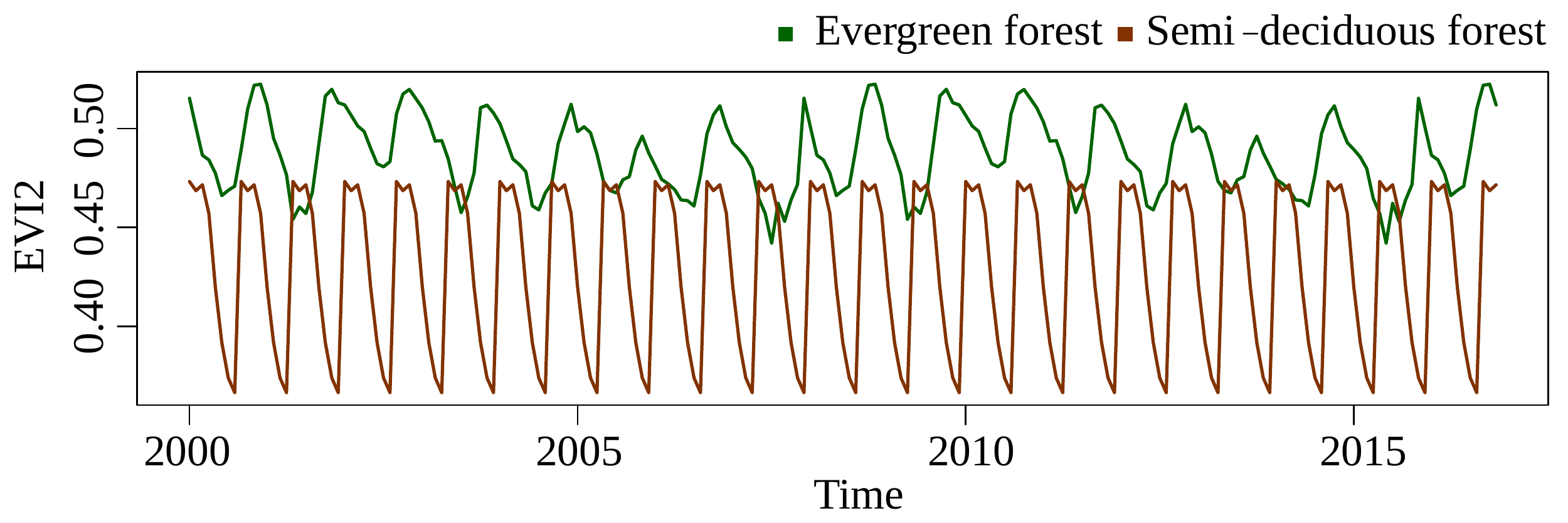}
\includegraphics[width=.49\columnwidth]{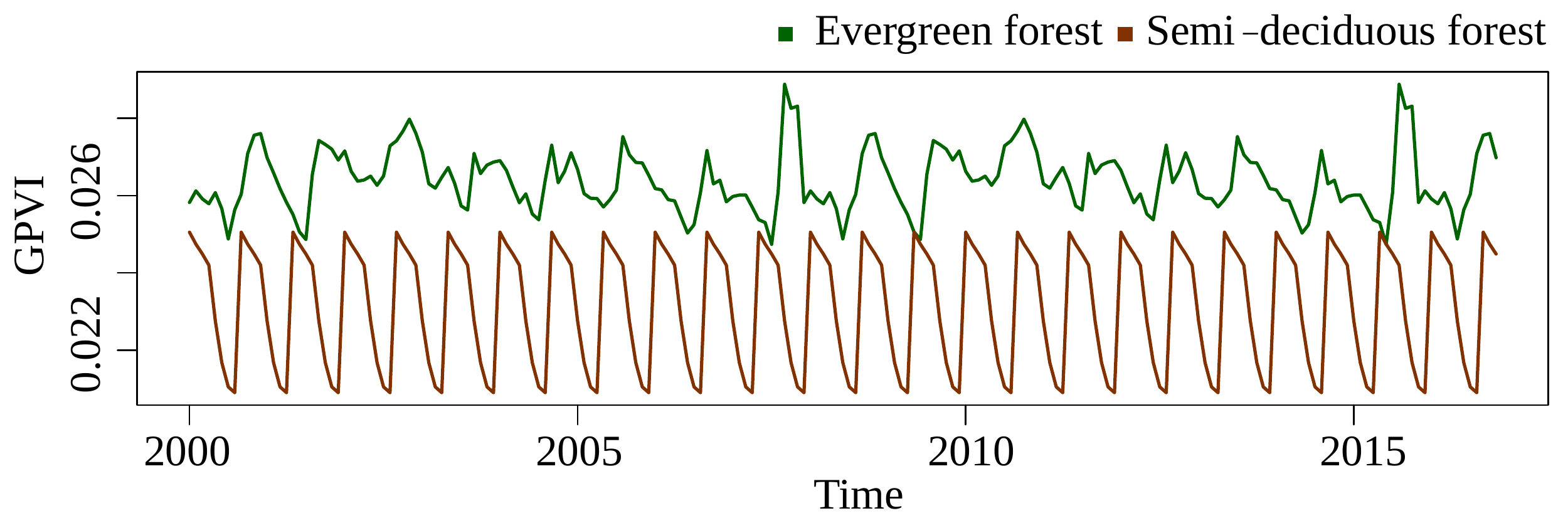}
\caption{Time series for evergreen forest/semi-deciduous forest discrimination (Landsat).}
\label{fig:ts-landsat-f}
\end{figure}

\begin{figure}[H]
\centering
\includegraphics[width=.49\columnwidth]{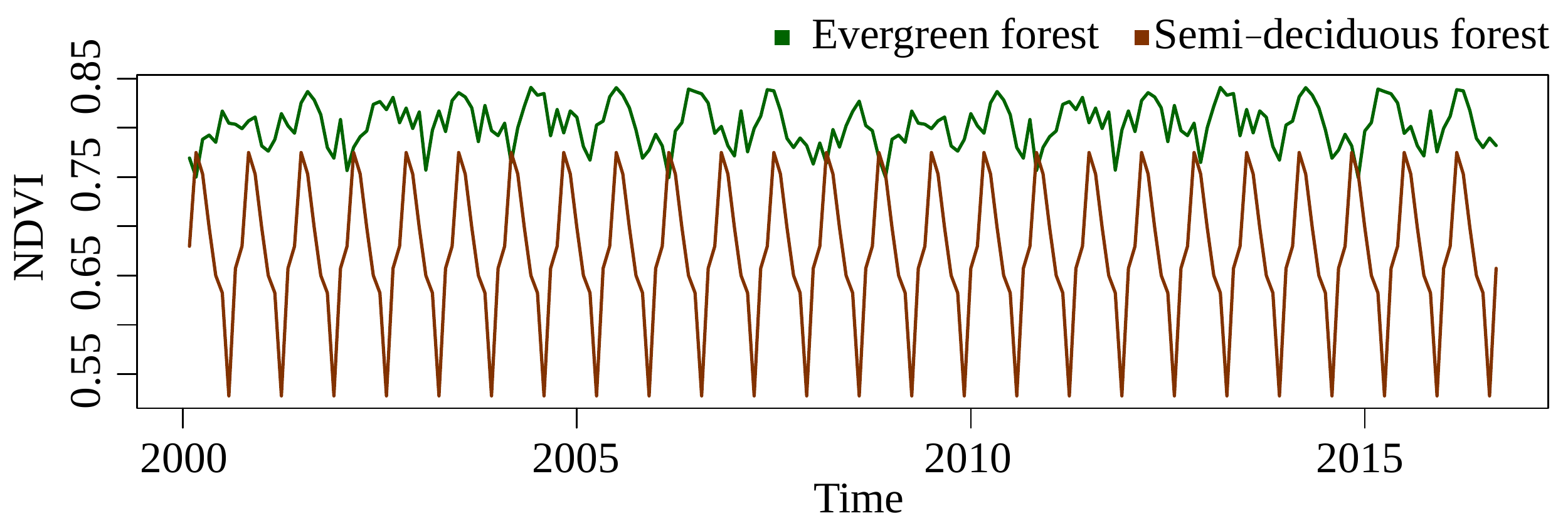}
\includegraphics[width=.49\columnwidth]{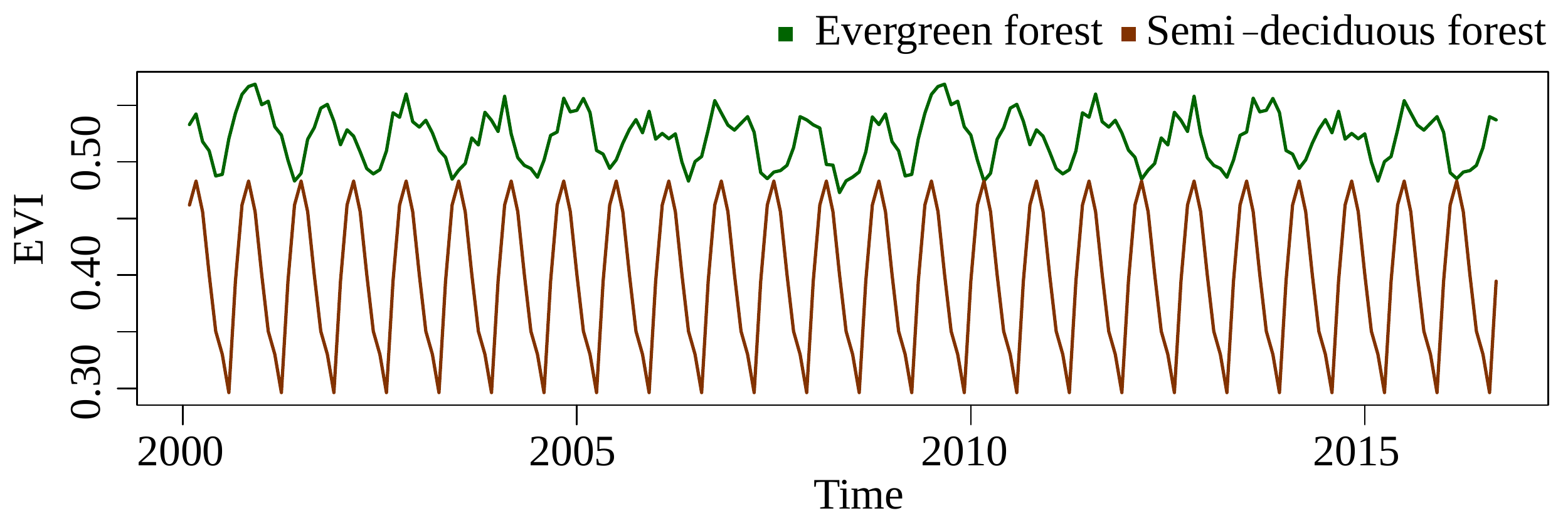}
\includegraphics[width=.49\columnwidth]{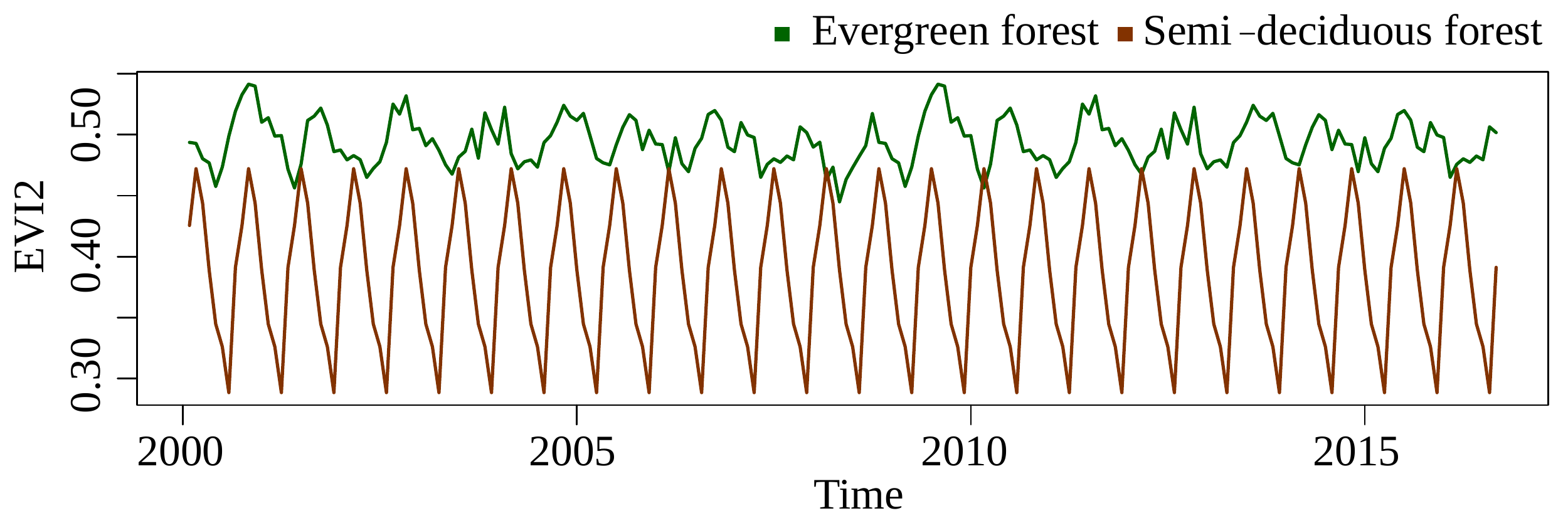}
\includegraphics[width=.49\columnwidth]{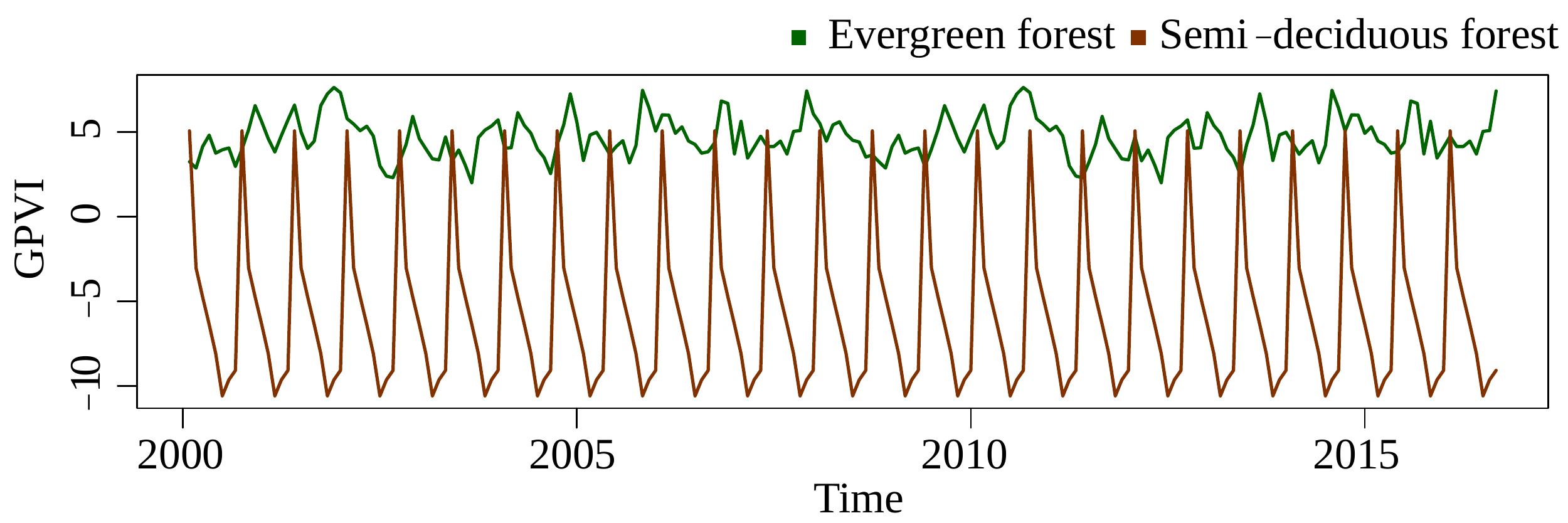}
\caption{Time series for evergreen forest/semi-deciduous forest discrimination (MODIS).}
\label{fig:ts-modis-f}
\end{figure}

The next classification problem corresponds to evergreen forest/semi-deciduous forest. Figures~\ref{fig:ts-landsat-f}~and~\ref{fig:ts-modis-f} show the time series for Landsat and MODIS, respectively. Although~the separation of GPVI is not as good as it was in the forest/savanna problem, it is still better than the baselines. Moreover, the~difference between sensors is also not as~evident.
\unskip

\begin{figure}[H]
\centering
\includegraphics[width=.49\columnwidth]{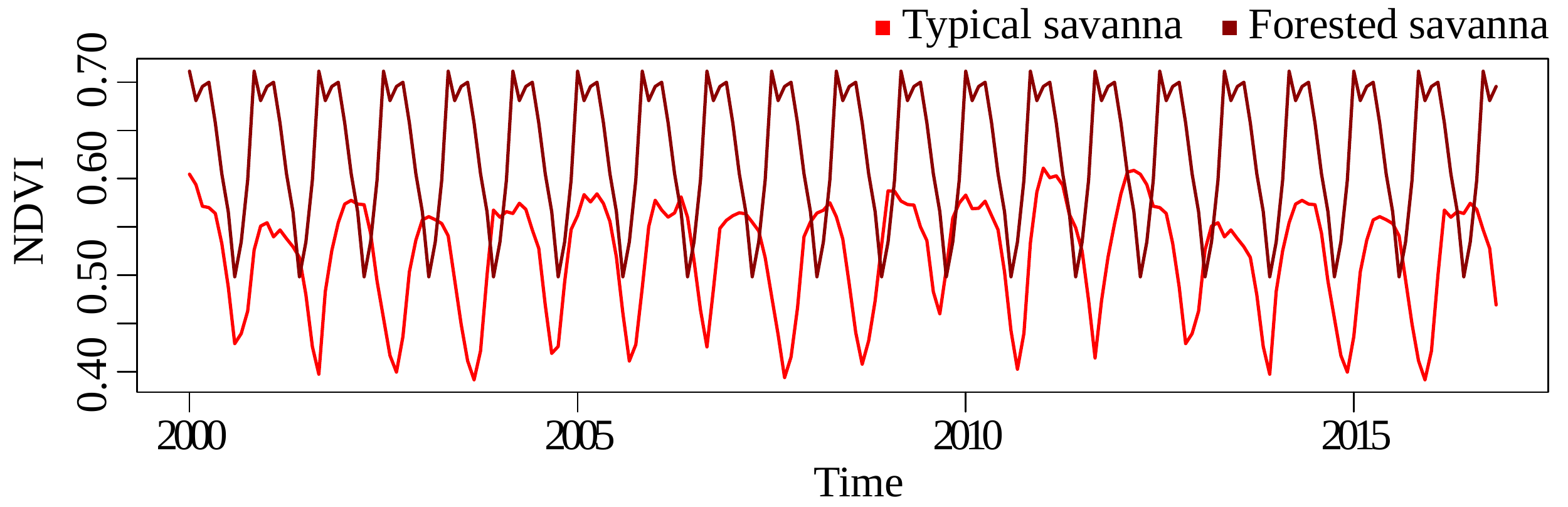}
\includegraphics[width=.49\columnwidth]{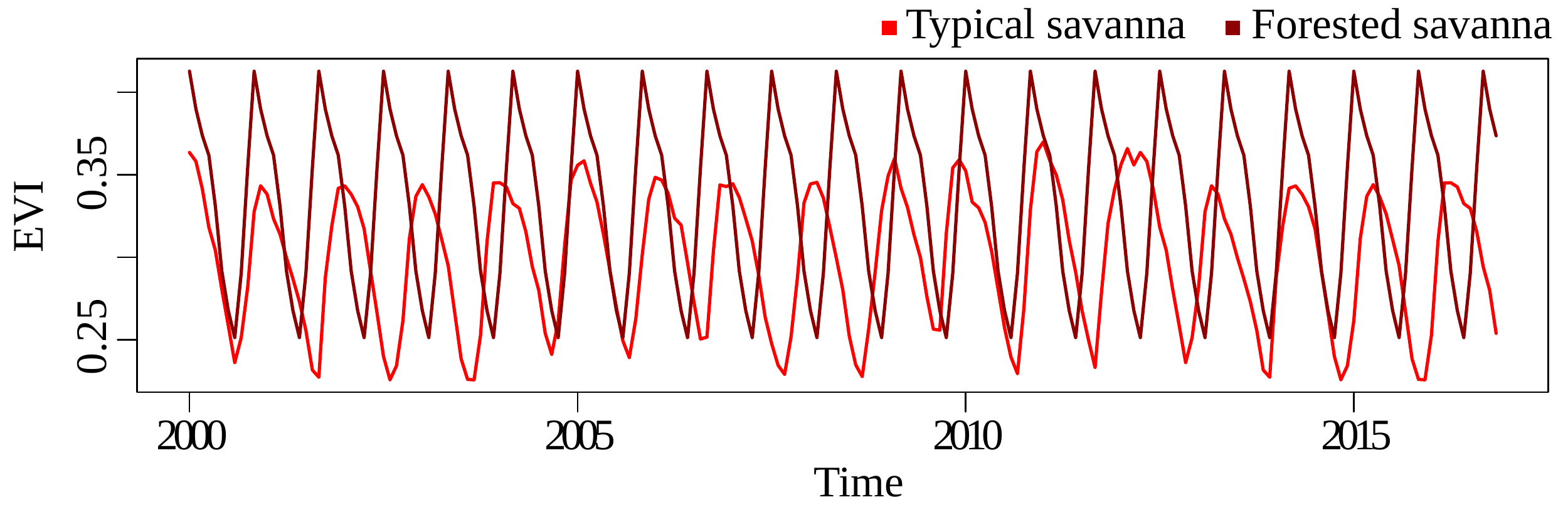}
\includegraphics[width=.49\columnwidth]{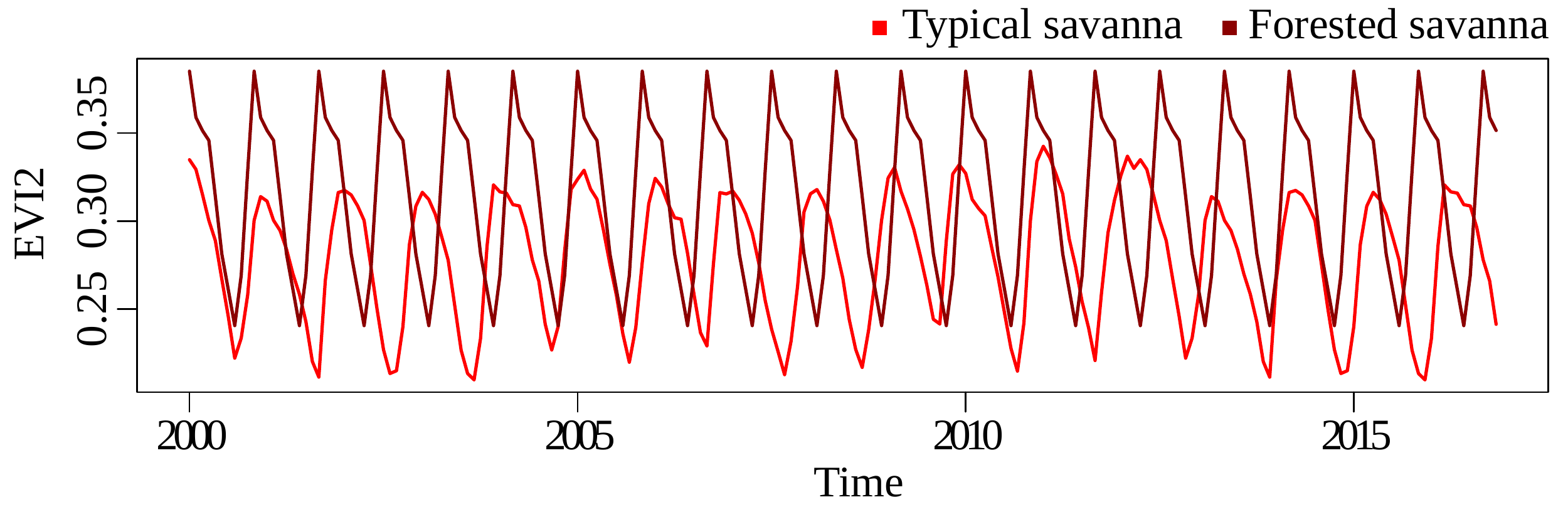}
\includegraphics[width=.49\columnwidth]{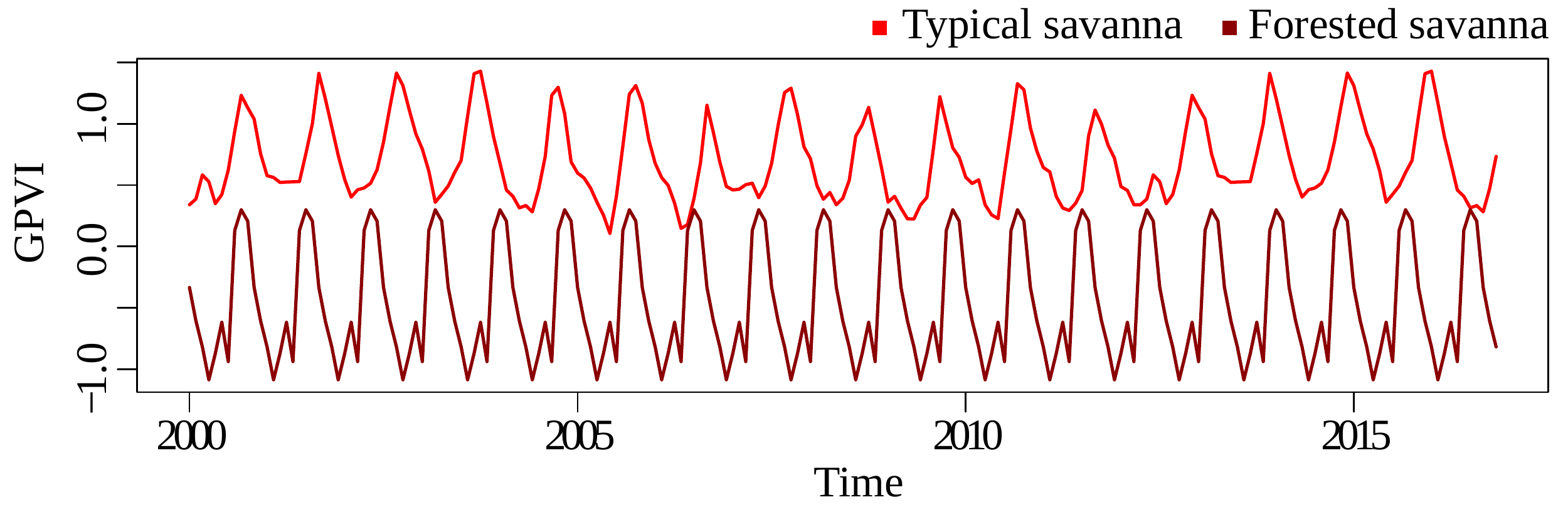}
\caption{Time series for typical savanna/forested savanna discrimination (Landsat).}
\label{fig:ts-landsat-s}
\end{figure}

\begin{figure}[H]
\centering
\includegraphics[width=.49\columnwidth]{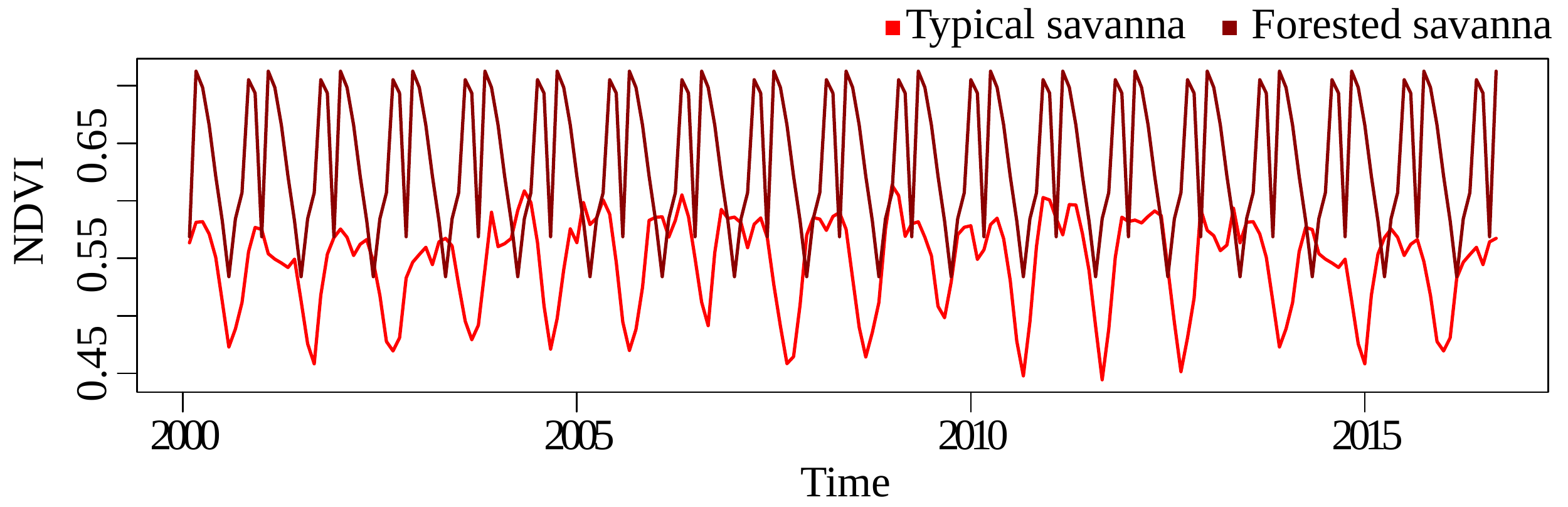}
\includegraphics[width=.49\columnwidth]{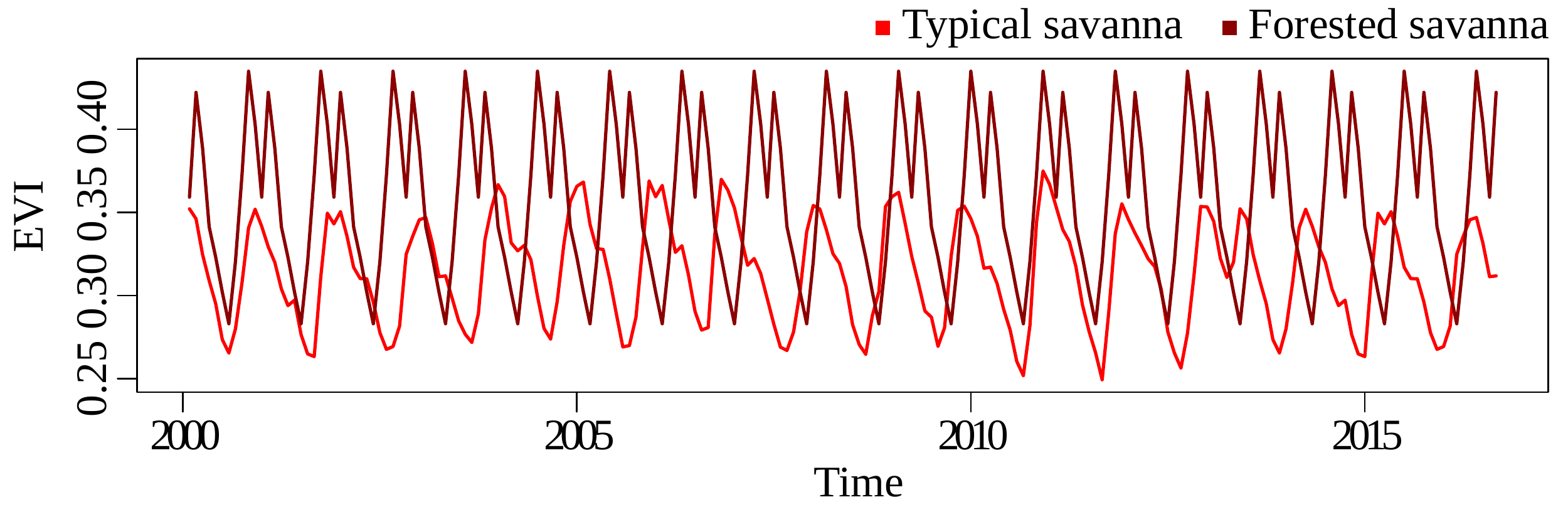}
\includegraphics[width=.49\columnwidth]{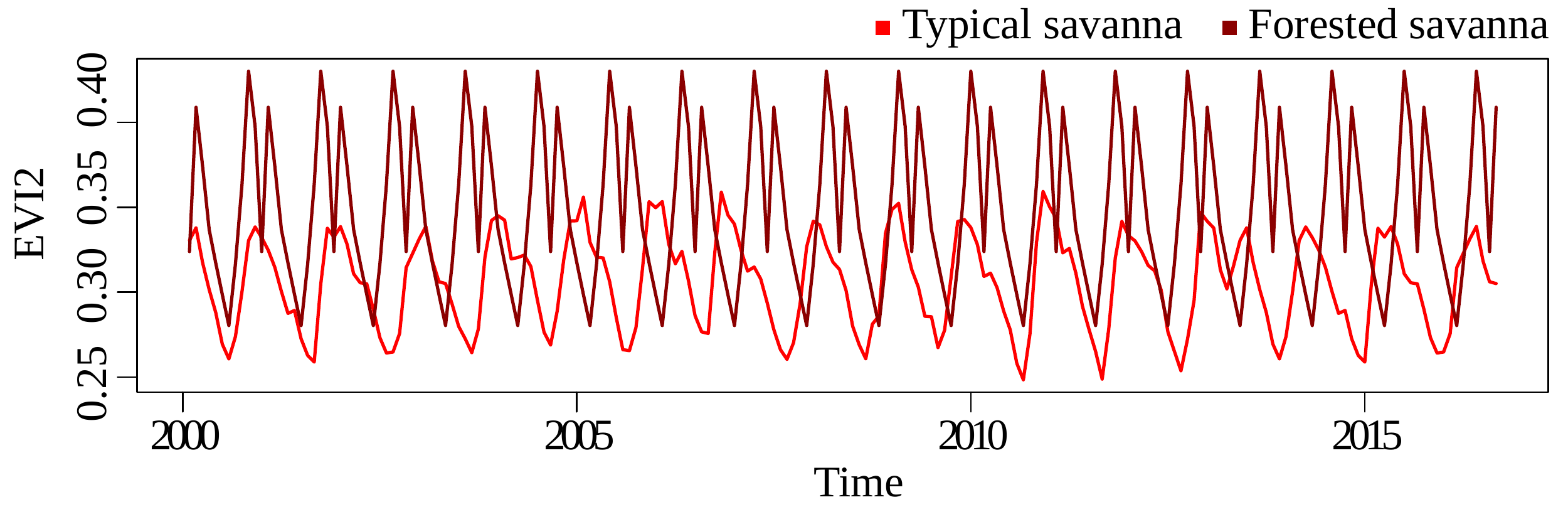}
\includegraphics[width=.49\columnwidth]{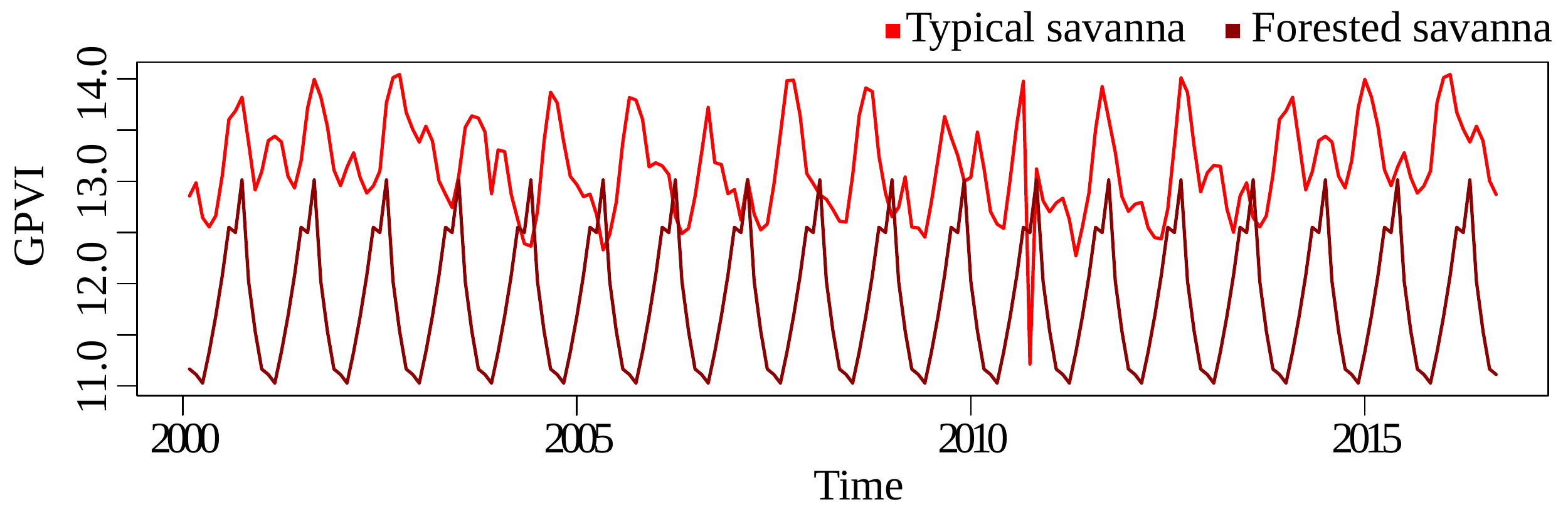}
\caption{Time series for typical savanna/forested savanna discrimination (MODIS).}
\label{fig:ts-modis-s}
\end{figure}

Figures~\ref{fig:ts-landsat-s}~and~\ref{fig:ts-modis-s} show the time series for Landsat and MODIS, respectively, for~the typical savanna/forested savanna savanna classification problem. The~separation of the classes for all indices was worse than the other classification problems, showing that discriminating between these two classes is a hard problem---at least with the available spectral~information.

\unskip

\unskip

\subsubsection{Time Series Classification~Performance}

Table~\ref{tab:ts-acc-fs} shows the normalized accuracy values that were obtained in the experiment corresponding to the forest/savanna time series discrimination for both sensors, as~well as the user's and producer's accuracy. The~Friedman test p-value associated with Landsat is $3.67\times 10^{-5}$, and~with MODIS was $7.28\times 10^{-2}$, showing that, for~both sensors, there was a significant statistical difference. However, the~corrected pairwise tests show that there is no significant difference between GPVI and NDVI for Landsat. For~MODIS, the~tests show no significant difference between GPVI and EVI/EVI2.

\begin{table}[H]

\centering
\caption{Normalized accuracies and user's and producer's accuracies per class for forest/savanna time series discrimination. $\blacktriangle$ means that GPVI was statistically superior to the corresponding baseline; $\blacktriangledown$,~which it was statistically inferior, and~$\bullet$, which both methods were~tied.}
\renewcommand{\arraystretch}{1.2}
\begin{tabular}{ccccccc}
\toprule
\multicolumn{1}{c}{} &
\multicolumn{1}{c}{} &
\multicolumn{2}{c}{\textbf{Forest}} & \multicolumn{2}{c}{\textbf{Savanna}} & \multicolumn{1}{c}{\textbf{Total}} \\ \midrule
\textbf{Sensor} & \textbf{Index} & \textbf{Producer} & \textbf{User} & \textbf{Producer} & \textbf{User} & \textbf{Accuracy} \\ \midrule
\multirow{4}{*}{Landsat} & NDVI & $98.23 \pm 2.20$ & $99.63 \pm 0.79$ & $99.26 \pm 0.64$ & $98.96 \pm 1.49$ & $\bullet~ 98.75 \pm 0.98$\\
& EVI & $94.71 \pm 2.80$ & $98.36 \pm 1.57$ & $98.48 \pm 1.14$ & $96.62 \pm 1.78$ & $\blacktriangle~96.60 \pm 1.05$\\
& EVI2 & $94.69 \pm 1.37$ & $98.34 \pm 1.98$ & $98.13 \pm 2.01$ & $97.22 \pm 1.59$ & $\blacktriangle~96.41 \pm 1.45$\\
& GPVI & $99.10 \pm 0.95$ & $99.62 \pm 0.81$ & $99.64 \pm 0.59$ & $99.48 \pm 0.92$ & $~~~99.37 \pm 0.36$\\ \midrule
\multirow{4}{*}{MODIS} & NDVI & $92.30 \pm 3.47$ & $91.72 \pm 3.74$ & $94.34 \pm 1.91$ & $98.68 \pm 2.57$ & $\bullet~93.32 \pm 1.60$ \\
& EVI & $95.73 \pm 1.86$ & $93.82 \pm 2.09$ & $95.13 \pm 2.14$ & $96.59 \pm 2.18$ & $\bullet~95.43 \pm 1.60$ \\
& EVI2 & $95.57 \pm 2.19$ & $93.58 \pm 2.81$ & $94.83 \pm 2.30$ & $95.28 \pm 2.51$ & $\bullet~95.20 \pm 1.42$ \\
& GPVI & $94.80 \pm 1.91$ & $92.37 \pm 3.23$ & $94.20 \pm 1.57$ & $93.39 \pm 2.65$ & $~~~94.50 \pm 1.46$ \\ \bottomrule
\end{tabular}
\label{tab:ts-acc-fs}
\end{table}

Table~\ref{tab:ts-acc-f} shows the normalized accuracy values that were obtained in the experiment corresponding to the evergreen forest/semi-deciduous forest time-series discrimination for both sensors, as~well as the user's and producer's accuracy. The~Friedman test p-value for Landsat was $0.2$ and, for~MODIS, was~$0.4$, showing that there is no significant difference among all of the indices in general. The~corrected pairwise tests show the same result. Unlike the forest/savanna problem, results using MODIS dataset shows better results than results with~Landsat.

\begin{table}[H]
\centering
\caption{Normalized accuracies and user's and producer's accuracies per class for evergreen forest/semi-deciduous forest time series discrimination. $\blacktriangle$ means that GPVI was statistically superior to the corresponding baseline; $\blacktriangledown$, which it was statistically inferior, and~$\bullet$, which both methods were~tied.}
\renewcommand{\arraystretch}{1.2}
\begin{tabular}{ccccccc}
\toprule
\multicolumn{1}{c}{} &
\multicolumn{1}{c}{} & \multicolumn{2}{c}{\textbf{Evergreen Forest}} & \multicolumn{2}{c}{\textbf{Semi-Deciduous Forest}} & \multicolumn{1}{c}{\textbf{Total}} \\ \midrule
\textbf{Sensor} & \textbf{Index} & \textbf{Producer} & \textbf{User} & \textbf{Producer} & \textbf{User} & \textbf{Accuracy} \\ \midrule
\multirow{4}{*}{Landsat} & NDVI & $93.72 \pm 4.09$ & $94.76 \pm 6.64$ & $23.69 \pm 24.84$ & $57.95 \pm 34.31$ & $\bullet~58.70 \pm 13.28$\\
& EVI & $94.59 \pm 3.68$ & $96.28 \pm 4.33$ & $40.10 \pm 31.82$ & $39.52 \pm 28.71$ & $\bullet~67.34 \pm 16.20$\\
& EVI2 & $94.25 \pm 3.00$ & $ 97.29 \pm 1.93$ & $41.50 \pm 25.46$ & $39.29 \pm 30.20$& $\bullet~67.88 \pm 12.56$\\
& GPVI & $96.31 \pm 3.71$ & $97.27 \pm 3.47$ & $54.00 \pm 27.43$ & $29.29 \pm 28.99$ & $~~~75.16 \pm 13.50$\\ \midrule
\multirow{4}{*}{MODIS} & NDVI & $95.16 \pm 2.41$ & $96.41 \pm 4.53$ & $48.67 \pm 32.48$ & $42.17 \pm 14.97$ & $\bullet~71.91 \pm 15.77$ \\
& EVI & $96.29 \pm 2.57$ & $98.09 \pm 1.26$ & $68.33 \pm 23.32$ & $44.50 \pm 17.20$ & $\bullet~82.31 \pm 11.16$ \\
& EVI2 & $96.10 \pm 2.30$ & $98.46 \pm 1.22$ & $68.33 \pm 30.24$ & $46.50 \pm 19.96$ & $\bullet~82.22 \pm 14.39$ \\
& GPVI & $97.19 \pm 2.45$ & $97.73 \pm 1.72$ & $65.45 \pm 31.74$ & $47.50 \pm 22.73$ & $~~~81.32 \pm 14.85$ \\ \bottomrule
\end{tabular}
\label{tab:ts-acc-f}
\end{table}

Table~\ref{tab:ts-acc-s} shows the normalized accuracy values that were obtained in the experiment corresponding to the typical savanna/forested savanna time series discrimination for both sensors, as~well as the user's and producer's accuracy. The~Friedman test p-value for Landsat is $6.67\times 10^{-4}$ and, for~MODIS, is $0.23$, showing that only for Landsat there was a significant difference. The~corrected pairwise tests evidence the superiority of GPVI as compared to EVI2 only for Landsat. For~this classification problem, both sensors obtained a similar~performance.

\begin{table}[H]
\centering
\caption{Normalized accuracies and user's and producer's accuracies per class for typical savanna/forested savanna time series discrimination. $\blacktriangle$ means that GPVI was statistically superior to the corresponding baseline; $\blacktriangledown$, which it was statistically inferior, and~$\bullet$, which both methods were~tied.}
\renewcommand{\arraystretch}{1.2}
\begin{tabular}{ccccccc}
\toprule
\multicolumn{1}{c}{} &
\multicolumn{1}{c}{} & \multicolumn{2}{c}{\textbf{Typical Savanna}} & \multicolumn{2}{c}{\textbf{Forested Savanna}} & \multicolumn{1}{c}{\textbf{Total}} \\ \midrule
\textbf{Sensor} & \textbf{Index} & \textbf{Producer} & \textbf{User} & \textbf{Producer} & \textbf{User} & \textbf{Accuracy} \\ \midrule
\multirow{4}{*}{Landsat} & NDVI & $94.55 \pm 2.55$ & $94.56 \pm 1.41$ & $20.83 \pm 19.71$ & $18.50 \pm 11.95$ & $\bullet~57.69 \pm 10.04$\\
& EVI & $94.26 \pm 2.73$ & $94.15 \pm 3.34$ & $14.86 \pm 12.38$ & $12.67 \pm 18.78$ & $\bullet~54.56 \pm 6.26$\\
& EVI2 & $93.32 \pm 2.66$ & $94.42 \pm 3.50$ & $0.00 \pm 0.00$ & $11.83 \pm 10.44$ & $\blacktriangle~ 46.66\pm 1.33$\\
& GPVI & $95.13 \pm 2.91$ & $92.38 \pm 2.03$ & $31.20 \pm 26.48$ & $12.67 \pm 16.24$ & $~~~63.16 \pm 12.66$\\ \midrule
\multirow{4}{*}{MODIS} & NDVI & $94.20 \pm 2.60$ & $95.25 \pm 1.63$ & $11.56 \pm 8.83$ & $5.33 \pm 8.64$ & $\bullet~52.88 \pm 5.13$ \\
& EVI & $94.26 \pm 2.62$ & $98.24 \pm 1.06$ & $26.42 \pm 31.94$ & $12.33 \pm 11.66$ & $\bullet~60.34 \pm 16.18$ \\
& EVI2 & $94.32 \pm 2.79$ & $98.12 \pm 1.35$ & $16.58 \pm 17.16$ & $5.67 \pm 9.17$ & $\bullet~55.45 \pm 9.35$ \\
& GPVI & $94.42 \pm 2.18$ & $94.23 \pm 2.93$ & $30.13 \pm 27.99$ & $18.33 \pm 20.68$ & $~~~62.28 \pm 13.80$ \\ \bottomrule
\end{tabular}
\label{tab:ts-acc-s}
\end{table}
\unskip

\subsection{Study on the Structure of the~GPVIs}
\label{subsec:bands}

To address~\ref{rq4}, we analyze the relevance that GPVI assigns to the bands and operators to attain a good classification, by~counting the most frequent formula elements in the top 10-indices obtained in the last population of the GP framework, for~the raw-pixel classification experiments (Sections~\ref{subsec:biome}~and~\ref{subsec:vegetation-type}). We compare the structure of the indices learned in Landsat and MODIS, and~the relevance of the associated bands for both sensors in order to check whetherthey match or~not.

\subsubsection{Experimental~Protocol}

As mentioned before, we considered the results of the raw-pixel classification experiments and, for~each classification problem, we extracted the 10 indices with the best performance that our GP framework found. While using the tree representation of the indices, we extracted all of the possible subtrees (one per inner node), corresponding to the subexpressions of a given formula. We~then count and rank the subexpressions, and~perform two analyses on this data: in the first one, we~build frequency histograms of all of the bands and compare the histograms yielded on the Landsat experiments, with~respect to the MODIS ones. In~the second analysis, we count how many times each subexpression is present in each experiment and show the 10 most frequent ones, when comparing both~sensors.

\subsubsection{Results}

We present the analysis on the structure of the GPVIs by analyzing the band and subexpression~relevance.

\paragraph{Band Relevance}

\begin{figure}[H]
\centering
\includegraphics[width=.6\columnwidth]{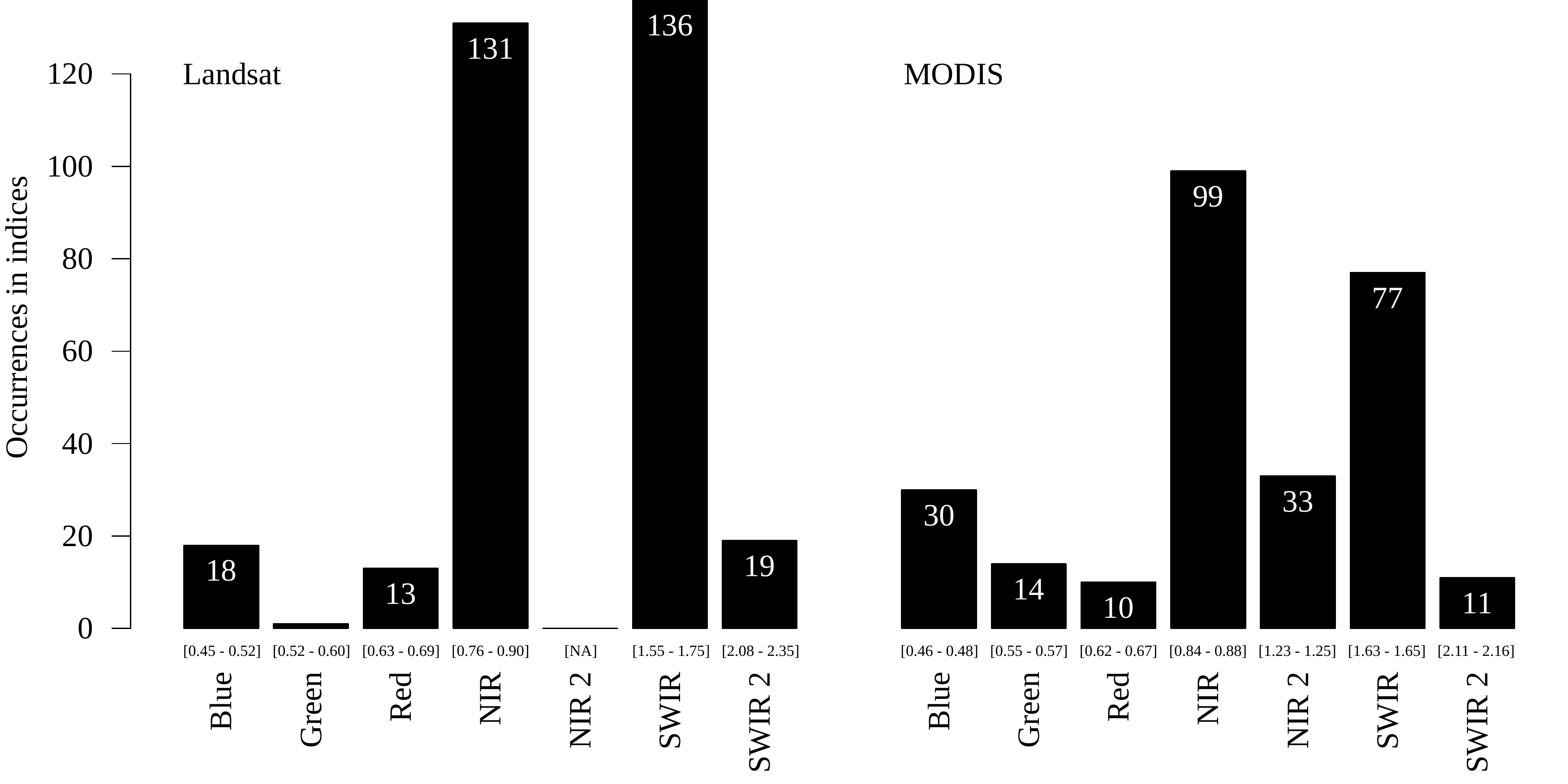}
\caption{Distribution of the frequencies of the spectral bands of the learned indices for forest/savanna~discrimination.}
\label{fig:lm-bands-relevance}
\end{figure}

\begin{figure}[H]
\centering
\includegraphics[width=.6\columnwidth]{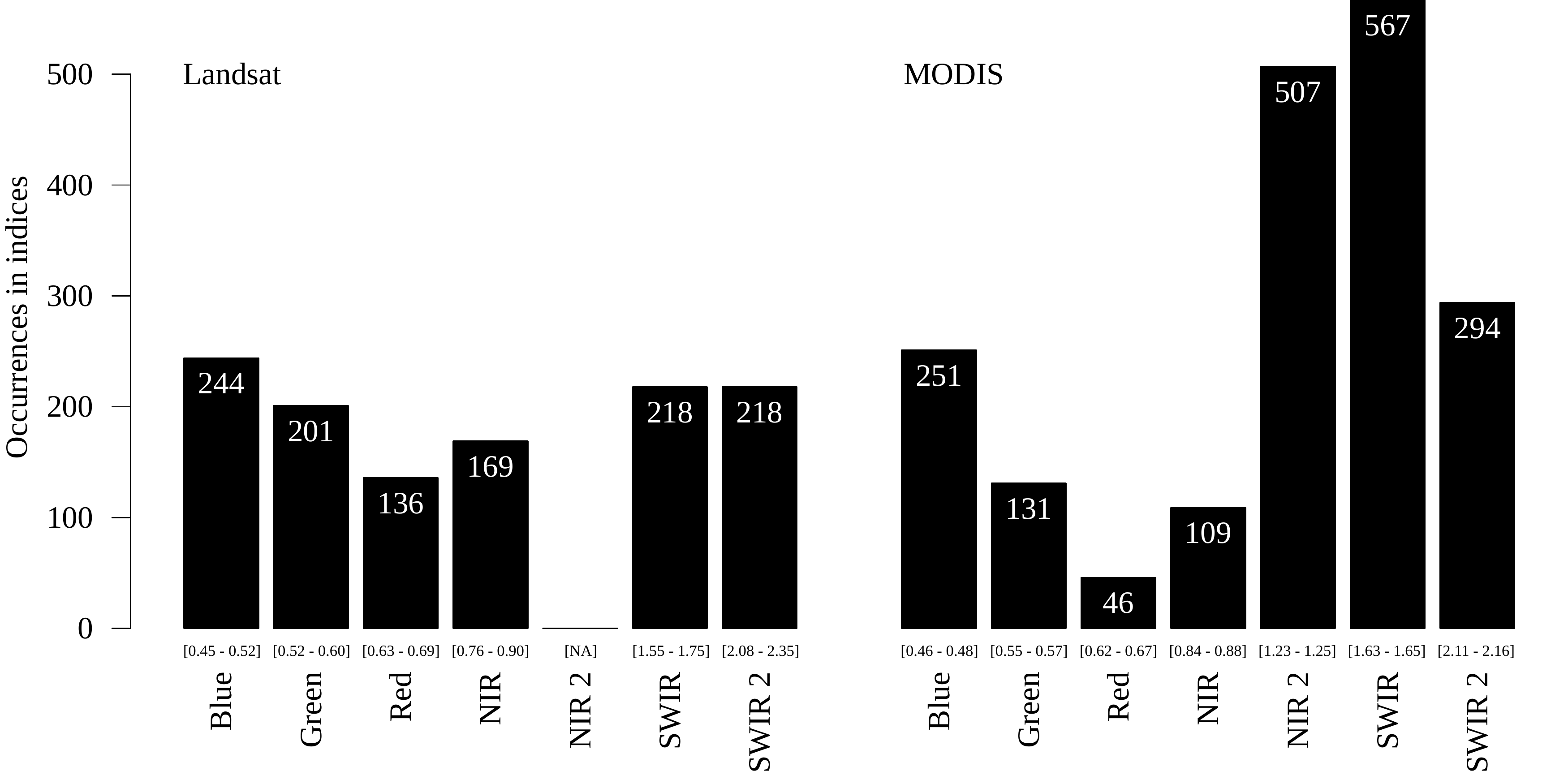}
\caption{Distribution of the frequencies of the spectral bands of the learned indices for evergreen forest/semi-deciduous forest~discrimination.}
\label{fig:lm-bands-relevance-f}
\end{figure}

\begin{figure}[H]
\centering
\includegraphics[width=.6\columnwidth]{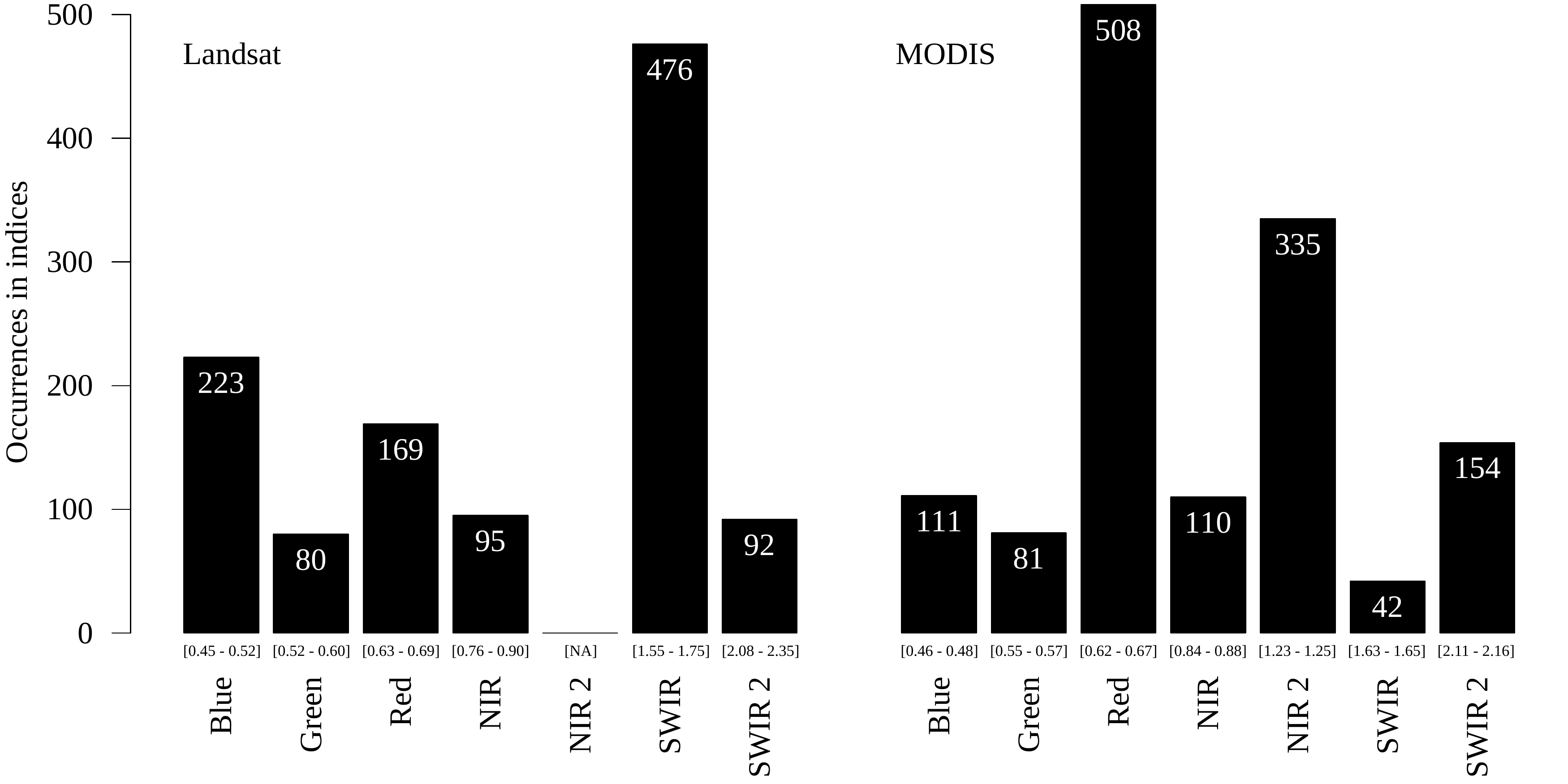}
\caption{Distribution of the frequencies of the spectral bands of the learned indices for typical savanna/forested savanna~discrimination.}
\label{fig:lm-bands-relevance-s}
\end{figure}

Figures~\ref{fig:lm-bands-relevance}--\ref{fig:lm-bands-relevance-s} show the number of occurrences of the bands with both sensors for, respectively, the~forest/savanna, evergreen forest/semi-deciduous forest, and~typical savanna/forested savanna classification~problems.

We analyzed the histograms of band frequencies for the learned indices of both sensor families (Landsat and MODIS) in order to identify similarities and differences regarding the use of specific bands in the GP-based indices. To~discriminate forests from savannas, there was an agreement between the two sensors, in~that the most frequent bands were NIR and SWIR. This differs from the baselines, which consider the Red channel instead of the SWIR one. In~fact, even the Blue channel (also used in EVI) was more frequently selected than the red one for both sensors. The~NIR2 channel was selected as frequently as the Blue for MODIS. Apparently, the~lack of this band did not give any disadvantage to the Landsat sensor to discriminate~correctly.

The classification problems by vegetation type did not attain a satisfactory agreement for band frequency between the sensors. The~separation by physiognomy between the same biome is clearly a harder~problem.

\unskip

\unskip

The NIR2 and SWIR bands for the MODIS sensor have clear relevance in the evergreen forest/semi-deciduous forest problem, while, for~Landsat, it is not prominent (Figure~\ref{fig:lm-bands-relevance-f}). The~only agreement between the sensors is that the Red and NIR bands, which are used in the baselines, are the least frequent. The~performance of GPVI for MODIS is about $11\%$ superior to Landsat (Table~\ref{tab:acc-f}), which~suggests that the NIR2 band has high relevance in characterizing these~classes.

The typical savanna/forested savanna problem presented no agreement between the sensors (Figure~\ref{fig:lm-bands-relevance-s}). For~example, the~SWIR band is the most frequent for Landsat and the least frequent for MODIS, and~the Red band, the~most frequent for MODIS, is not so frequent for Landsat. This was also the most difficult pair of classes to discriminate in previous~analyses.

\paragraph{Subexpression Relevance}

Table~\ref{tab:felements} shows the top-ten most frequent formula elements found in the indices for the three classification~problems.

\begin{table}[H]
\centering
\caption{List of the most frequent formula elements for the three classification~problems.}
\begin{tabular}{cccccccccc}
\toprule
\multicolumn{10}{c}{\textbf{forest/savanna}} \\ \midrule
\multicolumn{10}{c}{Landsat} \\ \midrule
\texttt{SWIR} & \texttt{NIR} & \texttt{-} & \texttt{\%} & \texttt{rlog} & \texttt{srt} & \texttt{NIR\%SWIR} & \texttt{*} & \texttt{+} & \texttt{rlog(NIR\%SWIR)} \\
136  & 131 & 127 & 120 & 104 & 92 & 83 & 55 & 29 & 22 \\ \midrule 
\multicolumn{10}{c}{MODIS} \\ \midrule
\texttt{NIR}  & \texttt{\%}  & \texttt{+} & \texttt{SWIR} & \texttt{-} & \texttt{srt} & \texttt{NIR2} & \texttt{*} & \texttt{NIR\%SWIR} & \texttt{Blue} \\
99   & 96  & 85 & 77 & 63 & 58 & 33 & 33 & 32 & 30 \\ \midrule
\multicolumn{10}{c}{\textbf{forest/semi-deciduous forest}} \\ \midrule
\multicolumn{10}{c}{Landsat} \\ \midrule
\texttt{\%} & \texttt{*} & \texttt{-} & \texttt{rlog} & \texttt{+} & \texttt{srt} & \texttt{Blue} & \texttt{SWIR} & \texttt{SWIR2} & \texttt{Green} \\
444 & 344 & 336 & 261 & 249 & 245 & 244 & 218 & 218 & 201 \\ \midrule 
\multicolumn{10}{c}{MODIS} \\ \midrule
\texttt{-}  & \texttt{SWIR}  & \texttt{\%} & \texttt{NIR2} & \texttt{+} & \texttt{*} & \texttt{srt} & \texttt{SWIR2} & \texttt{Blue} & \texttt{rlog} \\
622 & 470 & 436 & 425 & 356 & 279 & 276 & 224 & 189 & 181 \\\midrule
\multicolumn{10}{c}{\textbf{typical savanna/forested savanna}}\\\midrule
\multicolumn{10}{c}{Landsat} \\ \midrule
\texttt{rlog} & \texttt{SWIR} & \texttt{\%} & \texttt{*} & \texttt{srt} & \texttt{+} & \texttt{-} & \texttt{Blue} & \texttt{Red} & \texttt{NIR} \\
551 & 476 & 401 & 353 & 335 & 291 & 246 & 223 & 167 & 95 \\ \midrule
\multicolumn{10}{c}{MODIS} \\ \midrule
\texttt{srt}  & \texttt{Red} & \texttt{*} & \texttt{\%} & \texttt{+} & \texttt{-} & \texttt{NIR2} & \texttt{rlog} & \texttt{srt(Red)} & \texttt{SWIR2} \\
628 & 431 & 343 & 333 & 324 & 283 & 252 & 191 & 151 & 143 \\\midrule

\end{tabular}
\label{tab:felements}
\end{table}

Regarding the Landsat data, it can be seen that the most recurrent bands are NIR and SWIR, and~that the operations {SWIR--NIR and SWIR2 \% NIR appear various times. The~following is an example of a GPVI learned in the forest/savanna classification problem for~Landsat.\\

\noindent
\resizebox{\columnwidth}{!}{%
\begin{tabular}{|p{1\columnwidth}|}
\hline
\small\texttt{srt(Blue * rlog(SWIR - (NIR * SWIR2 - (SWIR - NIR)))) - ((NIR - (SWIR - (srt((srt((SWIR - (Blue * SWIR2 - (SWIR - NIR))) * Red) * rlog(rlog(rlog(srt(rlog(NIR) * ((Red * ((NIR - (rlog(rlog(Blue * SWIR2 - (SWIR2 \% NIR - ((SWIR - (srt(SWIR) - Blue * SWIR2)) * rlog(SWIR2 \% NIR))))) - Blue)) * Red)) - (SWIR - (Blue * SWIR2 - (SWIR - NIR))))))))) * rlog(SWIR2 \% NIR)) - ((NIR - (SWIR - NIR)) \% srt(SWIR))))) \% srt(SWIR))}\\
\hline
\end{tabular}}\\

For MODIS, the~NIR and SWIR bands are also the most recurrent, and~the operations NIR2 \% NIR and NIR \% SWIR appear various times. The~following is an example of a GPVI learned in the forest/savanna classification problem for~MODIS.\\

\noindent
\resizebox{\columnwidth}{!}{%
\begin{tabular}{|p{1\linewidth}|}
\hline
\small\texttt{srt(srt(NIR2 \% NIR)) \% ((NIR2 - srt(Green - (Green - (NIR2 + SWIR + (((Green \% SWIR2) * (NIR \% (NIR2 \% NIR))) * (((((NIR \% SWIR + 2 * Blue) * rlog(Red \% NIR)) - (srt(srt(NIR \% SWIR)) + 2 * Blue)) - (((NIR2 \% NIR) * (Red \% NIR)) * SWIR)) * ((((NIR \% (NIR2 \% NIR)) \% SWIR) + Blue) + Blue))))))) - (NIR \% SWIR + 2 * Blue))}\\
\hline
\end{tabular}}\\

For the classification of different physiognomies within the same biome, the~complexity of the learned indices increases significantly, and~a greater quantity of operations and constants appear in the~formulas.

\section{Discussion}
\label{sec:discussion}

The discussion about the results presented in Section~\ref{sec:results} is guided by the research questions defined in Section~\ref{sec:introduction}.
The performance of GPVI on the non-temporal-pixelwise forest/savanna classification problem~(\ref{rq1}) show a clear superiority of GPVI over NDVI, EVI, and~EVI2. This suggests that there exist complex interactions between the bands that characterize the class of a sample, regardless of its behavior through time. In~the baselines, savannas sometimes took high values that could be interpreted by the classification routine as forests, and~vice~versa (Figures~\ref{fig:ts-landsat-f-s} and~\ref{fig:ts-modis-f-s}). Thus, it is not possible to point out one area in a specific timestamp, measure its index, and~conclude its class. For~this reason, the~traditional classification benchmark generally involves temporal~information.

The relatively lower performance in the finer-scale classification problems, such as evergreen vs. semi-deciduous forest, and~typical savanna vs. forested savanna~(\ref{rq2}), probably responds to both technical and biological factors. Firstly, the~amount of data available for each physiognomical subcategory is significantly lower than for the biome classification problem (approximately half), and~the amount of data for each class is unbalanced, due to the predominance of evergreen forests over semi-deciduous forests, and~typical savannas over forested savannas in the study region. Those two factors are disadvantageous for most of the prediction tasks. However, discrimination by physiognomy within the same broader-scale biome type is, by~itself, clearly a more complex biological problem.
Moreover, while species composition could also be used to distinguish between these vegetation types, and~different species can present different spectral signatures~\cite{ALBERTON201462,ALMEIDA201449}, species composition can substantially overlap among the vegetation types considered here~\cite{Bueno2018}. All of these factors could have contributed to the limited results that were obtained for these specific classification~problems.

Unlike single-pixel classification, the~time series classification approach showed that the accuracy of GPVI was not significantly superior to the baselines~(\ref{rq3}). The~most relevant reason is the fact that GPVI was originally trained for pixel classification, and~the fitness function used drives the convergence of the algorithm to indices that minimize intra-class variance. This is clear and can be seen in Figures~\ref{fig:ts-landsat-f-s} and~\ref{fig:ts-modis-f-s}, in~which the GPVI series are not only better separated, but~also presented less variability through time. Part of the discriminative information in time series is, precisely, this~variation, and~the classification algorithm, which is based on DTW distance, took advantage of this variability in order to align/correct the~series.

The experiments we conducted regarding sensor invariance, which was orthogonal to all of the experiments involving the Tropical South America regions, showed a consensus between Landsat and MODIS about the relevance of the bands for biome classification~(\ref{rq4}). Despite the sensor used, the~most relevant channels are NIR and SWIR, which map, respectively, biomass content and soil moisture. This contrasts with indices traditionally used in vegetation studies in which the red channel has a leading role, meaning that, besides~leaf reflectance (mapped by the NIR channel), soil moisture is more relevant to discriminate these types of biomes, rather than vegetation slope. A~plausible explanation for this is that a closed canopy, as~observed in forests, reduces light incidence and prevents soil water losses, reducing evapotranspiration and allowing for more water to be retained in the~soil.

There was no consensus between the two sensors regarding the relevance of the bands for discriminating vegetation types within biomes. However, before~concluding that these problems are unsolvable for spectral indices, factors, like the reduced quantity of data for training, the~relatively low spectral resolution of the sensor, {or the fact that MODIS includes noisy bands due to their low spatial resolution,} must be considered. It is also important to notice that, unlike the forest/savanna problem, the~experiments in MODIS sensor attributed a high relevance to channel NIR 2 (which is not provided by the Landsat sensor) when discriminating vegetation types. In~both cases---evergreen forest vs. semi-deciduous forest and typical savanna vs. forested savanna---this channel was selected as the second most relevant. The~fact that this channel is not provided by one of the sensors probably affected the consensus between the two. However, this area of the spectrum is not recognized as relevant for land-cover analysis in other studies and it is rather used to determine cirrus cloud contamination. Thus, no sound and straightforward interpretation could be made in these results alone. \mbox{Nevertheless, we speculate} that this result could be related to the role of forest in climate regulation given that forests can increase cloud formation above them~\cite{zemp2017self}.

\section{Conclusions and Future~Work}
\label{sec:conclusions}

We introduced a Genetic-Programming-based framework for learning specialized vegetation indices from data, and~tested it for discriminating between tropical South American forests and savannas, as~well as the vegetation types existing within them. The~framework outputs an index, which we called GPVI, optimized to project the spectral data of each pixel into a space where they can be easily discriminated. In~the case of pixel-level discrimination, GPVI, when applied on each multispectral pixel to project them into a one-dimensional space, outperformed traditional indices, such as NDVI, EVI, and~EVI2, showing that it is possible to describe discriminative patterns of the land coverage, without~the use of entire time-series. Based on the same GPVI training procedure, we~also tested the index for time-series classification and obtained a performance equivalent to baseline indices, despite the fact that the index was designed for single time stamp discrimination. Our index also outperformed traditional ones at discriminating between vegetation types within biomes, despite the high complexity involved in this specific problem. We speculate that the GPVI performance for this task could be highly improved with larger data availability and higher spectral resolution. When~considering that biomes and vegetation types within biomes are dynamic and they can be replaced by one another over time in response to natural and anthropogenic disturbances, our~approach is a promising framework for understanding how biomes will respond to present and future human induced global change~drivers.

{Our GP framework is also effective on hyperspectral data, with~no further changes on its configuration. Experiments supporting this claim have already been published~\cite{hernandez2016,hernandez2017}, in~which the GP framework was tested on hyperspectral multi-class datasets, which not only contain vegetation areas, but~also other types of land cover categories, and~consider more than two classes. This demonstrates the robustness of our method with respect to the number and variety of bands, implying that GP is an approach that can naturally select and combine bands, regardless of heterogeneous setups, such as spatial/spectral resolution, type of land cover, and~number of classes.}

As future work, one promising approach can be to explicitly express the GP framework as multi-objective, as~has been done in previous works~\cite{Nag2016, liddle2010multi, bleuler2001multiobjective, shao2013feature, rodriguez2004identifying, tay2008evolving, liang2019figure}; this way, the~relative importance between inter-class and intra-class distance of the examples can also be optimized, and~novel objectives (\mbox{e.g., precision and} recall) can be~considered.

\vspace{6pt} 



\authorcontributions{Conceptualization, J.F.H.A. and R.d.S.T.; methodology, J.F.H.A. and R.d.S.T.; software, J.F.H.A.; validation, R.S.O., M.H. and J.A.d.S.; formal analysis, J.F.H.A.; investigation, J.F.H.A. and R.d.S.T.; resources, R.S.O. and M.H.; data curation, J.F.H.A.; writing---original draft preparation, J.F.H.A. and R.d.S.T.; writing---review~and editing, R.S.O., M.H. and J.A.d.S.; visualization, J.F.H.A.; supervision, R.d.S.T.; project~administration, R.d.S.T.; funding acquisition, R.d.S.T. All authors have read and agreed to the published version of the~manuscript.}


\funding{This research was funded by the Coordena\c c\~ao de Aperfei\c coamento de Pessoal de N\'ivel Superior---Brasil~(CAPES), grant \#88881.145912/2017-01---Finance Code 001, Conselho Nacional de Desenvolvimento Cient\'ifico e Tecnol\'ogico (CNPq), grants \#307560/2016-3 and \#132847/2015-9, Funda\c c\~ao de Amparo à Pesquisa do Estado de S\~ao Paulo (FAPESP) (grants \#2018/06918-3, \#2017/12646-3, \#2016/26170-8, and~\#2014/12236-1) and the FAPESP-Microsoft Virtual Institute (grants \#2016/08085-3, \#2015/02105-0, \#2014/50715-9, \#2013/50169-1, and~\#2013/50155-0).}

\acknowledgments{The authors thank   Vinicius de L. Dantas for making available the ground-truth data of Tropical South America used in our study and for the overall support, and~to Alexandre Esteves Almeida for his contribution on data acquisition/curation.}

\conflictsofinterest{The authors declare no conflict of~interest.}
\vspace{-6pt}




\reftitle{References}

\end{document}